\documentclass[journal]{IEEEtran}
\usepackage{cite}
\usepackage{amsmath,amssymb,amsfonts}
\usepackage{algorithmic}
\usepackage[linesnumbered,ruled]{algorithm2e}
\usepackage{graphicx}
\usepackage{textcomp}
\usepackage{pifont}
\usepackage{xcolor}
\usepackage{soul}
\usepackage{xcolor}
\usepackage{cancel}

\def\BibTeX{{\rm B\kern-.05em{\sc i\kern-.025em b}\kern-.08em
    T\kern-.1667em\lower.7ex\hbox{E}\kern-.125emX}}
\begin{document}

\title{CoDrone: Autonomous Drone Navigation Assisted by Edge and Cloud Foundation Models}

\author{


\author{Pengyu Chen, Tao Ouyang, Ke Luo, Weijie Hong and Xu Chen, ~\IEEEmembership{Senior Member,~IEEE}
\thanks{Corresponding author: Xu Chen.}
\thanks{Pengyu Chen and Xu Chen are with the School of Computer Science and Engineering, Sun Yat-sen University, Guangzhou, Guangdong 510006, China (e-mail: chenpy77@mail2.sysu.edu.cn;   chenxu35@mail.sysu.edu.cn).}
\thanks{Tao Ouyang was with the School of Computer Science and Engineering, Sun Yat-sen University, Guangzhou, Guangdong, China, and is now with the School of Electronic Information, Central South University, Changsha, Hunan 410004, China (e-mail: ouyangtao@csu.edu.cn)}
\thanks{Ke Luo is with Pengcheng Laboratory, Shenzhen, Guangdong 518000, China (e-mail:  luok@pcl.ac.cn).}
\thanks{Weijie Hong is with Shenzhen Smart City Communications Co., Ltd., China (e-mail: hongweijie@smartcitysz.com)}
}}

\maketitle

\begin{abstract}
Autonomous navigation for Unmanned Aerial Vehicles (UAVs) presents significant challenges due to the limited onboard computational resources, which often restrict deployed deep neural networks to shallow architectures incapable of handling complex environments. Additionally, offloading tasks to remote edge servers introduces high latency, creating an inherent trade-off in system design. To address these limitations, we propose CoDrone—the first cloud-edge-end collaborative computing framework that integrates foundation models into autonomous UAV cruising scenarios—effectively leveraging foundation models to enhance the performance of resource-constrained unmanned aerial vehicle platforms.
To reduce both onboard computation and data transmission overhead, CoDrone employs grayscale imagery for the navigation model. When enhanced environmental perception is required, CoDrone leverages the edge-assisted foundation model Depth Anything V2 for depth estimation and introduces a novel, one-dimensional occupancy grid–based navigation method—enabling fine-grained scene understanding while significantly advancing the efficiency and representational simplicity of autonomous navigation.
A key component of CoDrone is a Deep Reinforcement Learning (DRL)-based neural scheduler that seamlessly integrates depth estimation with autonomous navigation decisions, enabling real-time adaptation to dynamic environments. Furthermore, the framework introduces a UAV-specific vision language interaction module, which incorporates domain-tailored low-level flight primitives to enable effective interaction between the cloud foundation model, the Vision Language model, and the UAV. 
The introduction of VLM enhances open-set reasoning capabilities in complex and previously unseen scenarios. We implement a prototype of CoDrone and conduct extensive evaluations in the AirSim simulation environment. Experimental results demonstrate that CoDrone significantly outperforms baseline methods under varying flight speeds and network conditions, achieving a 40\% increase in average flight distance and a 5\% improvement in average Quality of Navigation.
\end{abstract}

\begin{IEEEkeywords}
Autonomous navigation, edge computing, depth estimation model,
vision language model, dynamic offloading, 
\end{IEEEkeywords}

\section{Introduction}
The low-altitude economy, defined as an economic ecosystem leveraging low-altitude airspace (typically below 1,000 meters above ground level) and advanced aerial technologies to enable new value-added services, has emerged as a pivotal driver of global digital and industrial transformation. Breakthroughs in electrification, artificial intelligence (AI), and communication technologies—including the proliferation of electric vertical takeoff and landing (eVTOL) aircraft and intelligent unmanned aerial systems (UAS)—are transforming traditional aviation into an integrated component of urban mobility, logistics, agriculture, and public services\cite{cheng2024networkedisaclowaltitudeeconomy}, with projected global market value reaching \$400 billion by 2030\cite{DeliveryDonesMarketSize}. These low-altitude systems critically enable ubiquitous intelligence across diverse domains, ranging from daily activities to urban infrastructure development.



\begin{figure}[htbp]
    \centering
    \begin{minipage}{.99\linewidth}
        \includegraphics[width=\linewidth]{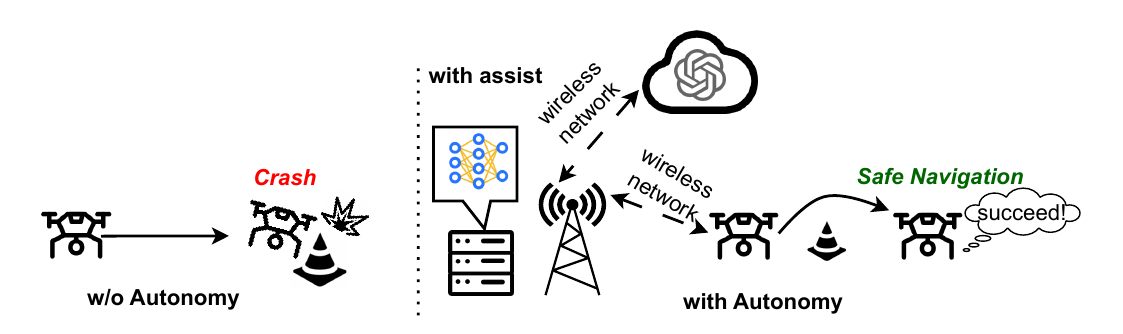}
        \caption{Autonomy and assist ensure safe navigation.}
        \label{fig:autonomy}
    \end{minipage}
\end{figure}
Small and medium-sized autonomous unmanned aerial vehicles (UAVs), serving as primary platforms in low-altitude IoT systems, integrate advanced sensing and intelligent capabilities. They are gradually penetrating various applications - including precision agriculture \cite{kurkute2018drones}, smart manufacturing, drone delivery (e.g., JD.com's logistics network), remote monitoring \cite{kaleem2018amateur}, and disaster response \cite{mishra2020drone} - primarily functioning as highly mobile image acquisition systems for real-time data analysis. Autonomous navigation, essential for obstacle avoidance without human intervention, is critical for mission success. As illustrated in Figure \ref{fig:autonomy}, effective onboard or edge-assisted autonomous cruise enables dynamic trajectory planning, obstacle evasion, and rapid mission completion in large-scale, complex terrains. Mission failure typically occurs due to collisions with obstacles or falls. Most UAVs implement autonomous cruising via Deep Neural Networks (DNNs), leveraging their powerful representation capacity to process visual inputs\cite{DistillingTinyForUAVS, Sim-to-Real, AutonomousNavigationofUAVs, drones7040245}. Specifically, the DNN model computes optimal real-time deflection angles and flight speeds based on the collected imagery, guiding immediate trajectory adjustments.
Crucially, reliable autonomous operation demands both high-quality image inputs and computationally intensive models, typically requiring larger-scale DNNs. 
However, the adoption of advanced sensors (e.g., RGB-D cameras, LiDAR, radar, ultrasound devices) is constrained by high manufacturing and maintenance costs in resource-limited UAVs. These limitations, coupled with the computational burden of large DNNs, create a significant challenge: achieving robust autonomous navigation in complex environments using only monocular cameras (i.e., a cost-effective sensor) to overcome unknown terrain hazards.

Previous studies have extensively investigated DNNs and deep reinforcement learning (DRL) for enhancing UAV autonomous navigation precision, with Wang \cite{AutonomousNavigationofUAVs} and Kalidas \cite{drones7040245} pioneering end-to-end DRL approaches that achieved significant performance gains. However, these frameworks fundamentally neglect critical operational constraints, including stringent onboard computational limitations and power budgets, which induce thermal constraints and energy depletion that severely compromise mission longevity. While the adaDrone framework \cite{chen2022adadrone} addresses these limitations through computational offloading to edge resources, it introduces dependency on stable edge connectivity and incurs substantial communication overhead. Peng et al. \cite{Uav-assisted} proposed a DRL-based path planning algorithm to optimize UAV trajectories in edge-assisted networks, but their reliance on lightweight DNNs significantly degrades robustness in complex environments, resulting in exponentially increasing collision susceptibility as environmental complexity escalates. Although foundation models, particularly Vision-Language Models (VLMs) \cite{saxena2025uavvlnendtoendvisionlanguage}, have demonstrated promising potential for enhancing cognitive navigation capabilities, their substantial inference latency (typically exceeding 200 ms) fundamentally conflicts with the stringent real-time requirements of UAV operations. Consequently, this research proposes a novel adaptive navigation framework that successfully integrates the cognitive advantages of foundation models with UAV-specific operational constraints, addressing the critical limitations of current approaches while meeting the rigorous real-time performance requirements essential for practical UAV deployment in low-altitude IoT systems.


However, achieving the aforementioned ideal paradigm in practice presents several significant challenges. First, conventional DNN-based navigation methods exhibit increasing limitations when operating in complex, dynamic environments. These methods typically rely on real-time visual inputs to generate low-level steering commands for obstacle avoidance. Nevertheless, in scenarios where the field of view is severely occluded, such as when encountering walls or failing to make timely decisions near obstacles, shallow DNN models frequently exhibit inadequate inference reliability under such conditions, which in turn increases the likelihood of suboptimal navigation outcomes or even task-level degradation. Consequently, it is crucial to enhance the fine-grained perceptual capabilities of existing approaches by leveraging emerging foundational models, ideally extending to semantic-level environmental understanding, thereby improving decision-making robustness.
Secondly, in real-world scenarios, network conditions exhibit frequent fluctuations, and operational contexts undergo rapid changes. Solely relying on onboard or edge-assisted computing may lead to performance degradation caused by latency or even navigation failures. Consequently, in uncertain environments, effectively integrating onboard navigation modules with auxiliary foundational models while ensuring real-time performance, safety, and adaptability, remains a challenging research problem that warrants further investigation.
Finally, extensive empirical evidence\cite{liu2023aerialvlnvisionandlanguagenavigationuavs} has demonstrated the efficacy of Vision Language Models (VLMs) in addressing diverse scenarios for autonomous UAV navigation. However, the substantial computational and storage requirements of these models present significant barriers to their direct deployment on edge server, thereby hindering the realization of low-latency inference capabilities. Consequently, there exists an urgent research imperative to investigate hybrid architectures that synergistically integrate the superior semantic understanding capabilities of cloud-based VLMs with conventional edge-assisted frameworks. Such an approach must rigorously balance latency constraints, system security, and computational efficiency to effectively overcome the inherent limitations of lightweight DNN models in complex navigation tasks.

To address the aforementioned challenges, we propose CoDrone, a cloud-edge-end cooperative framework for autonomous drone inspection tailored to dynamic edge computing environments. To tackle the first challenge related to onboard computational constraints, CoDrone reduces inference overhead by utilizing grayscale images instead of more resource-intensive RGB inputs during navigation processing. Furthermore, an edge foundation model, the depth estimation model Depth Anything V2 \cite{depth_anything_v2}, is uniformly deployed on the edge server. When necessary, the extracted depth information is transformed into a depth occupancy grid map, thereby enhancing fine-grained environmental perception.
In highly complex or previously unseen environments, where conventional methods may fail, CoDrone innovatively integrates a cloud foundation model, the Vision Language Model(VLM), to provide high-level semantic guidance. This integration allows the drone to resolve perceptual ambiguities and handle previously unseen scenarios more reliably, thereby enhancing the robustness and environmental adaptability of the navigation system. To tackle the second challenge, CoDrone introduces the Quality of Navigation (QoN) metric \cite{chen2022adadrone} to quantitatively assess navigation performance. Building upon this metric, CoDrone incorporates a DRL-based neural scheduler, which is carefully designed to learn and execute optimal scheduling strategies. By further integrating adaptive policies, the scheduler supports high-quality task offloading and foundational model-assisted navigation, thereby maximizing overall navigation performance. To address the third challenge, CoDrone introduces a UAV-specific vision-language interaction framework, which consists of two key components: a vision language model Module and an Invocation Strategy Module. 
The Vision Language Module is designed with customized, low-level flight primitives tailored to UAV operations. These primitives are encapsulated as callable functions within the system architecture, enabling the VLM to invoke them based on its semantic understanding of the environment and high-level mission instructions.  
Complementing this, the Invocation Strategy Module determines the optimal timing for engaging the cloud-deployed VLM by analyzing the drone’s historical flight data in conjunction with the output from the Neural Scheduler. This strategic invocation mechanism enables coordinated interaction with cloud-based VLM services, allowing the system to leverage semantic reasoning capabilities when necessary while preserving real-time responsiveness and operational autonomy.

\begin{itemize}
    \item To the best of our knowledge, this paper is the first to exploit the cloud-edge-end collaborative autonomous drone navigation utilizing edge and cloud foundation models. By leveraging the mutual benefits between DNNs and foundational models(Depth Estimation Model and Vision-Language Model), we synergize these tasks to achieve efficient autonomous navigation for more intelligent drones.
    \item To further reduce onboard computational load and data transmission latency, we utilize grayscale images and have specifically redesigned a lightweight DNN for autonomous drone navigation. 
    Additionally, we have devised a DRL-based neural scheduler and invocation strategy for adaptive optimizations that well leverage the assistance of both edge and cloud foundation models. 
    \item We implement a prototype of CoDrone based on the popular AirSim simulator. Extensive experiments under various scenarios validate CoDrone’s effectiveness. Results show that CoDrone outperforms baseline methods across different flight speeds and network conditions, achieving 40\% longer average flight distance and a 5\% improvement in average QoN.
\end{itemize}
The rest of this paper is organized as follows. Sec. \ref{sec:related_works} reviews the related works. Sec. \ref{sec:background_motivation} briefly reviews the autonomous navigation of drones. Sec. \ref{sec:codrone} overviews the system design of CoDrone and introduces how the different modules benefit each other. \ref{sec:implementation} describes the implementation of our prototype and simulation environment. Sec. \ref{sec:performance_eval} provides the evaluation results and Sec. \ref{sec:conclusion} concludes.

\section{Related Work}
\label{sec:related_works}
\textbf{Autonomous Drone Navigation.} Academic research in autonomous navigation of autonomous drones has huge interests in developing deep learning methods that enable drones to navigate and explore their environment using lightweight, low-cost, and low-power consuming neural networks \cite{gandhi2017learning, cao2019optimal, foehn2022agilicious}. These methods often leverage technologies such as depth estimation\cite{hong2021semantically} and semantic segmentation\cite{mandel2020towards} to enable drones to perceive and navigate their surroundings. However, the non-reconcilable conflict between the robustness of the perception module and the computational cost limits the development of those methods. Incorporating accompanying inspection analysis, CoDrone leverages extracted semantic information to augment the drone's navigation capabilities.

\textbf{Edge-assisted Drone Vision.} The emerging edge intelligence paradigm\cite{janssen2023supporting} provides drones an opportunity to offload computationally intensive tasks like object detection, image segmentation, and even real-time obstacle avoidance to edge servers. This frees up precious onboard resources for essential drone functions like flight control and sensor data acquisition\cite{wu2021deep}.
Serving those compute-intensive analysis tasks on the edge server with the UAV-captured images has garnered significant attention within the academic community\cite{nigade2022jellyfish, hao2023reaching}. A gap exists in existing research regarding the coordination and interplay between analysis tasks during inspection and the autonomous navigation of drones. Specifically, these studies often do not address the interdependencies between these tasks while CoDrone utilizes the relationship.

\textbf{Multi-Task Execution in Edge Intelligence System.}
Processing multiple tasks simultaneously in an edge intelligence system is pervasive because real-time applications always need to execute a set of DL components. Synergizing those tasks is essential to enhance the overall system performance. Some works\cite{ling2021rt,kong2023accumo} focus on coordinating multiple DNNs on the onboard computational platform by exploiting the software-hardware co-optimization. Several research papers\cite{cao2019optimal,zhang2023octopus} address the challenges of coordinating tasks across heterogeneous devices, managing limited resources, and ensuring efficient and secure task execution by leveraging deep reinforcement learning-based scheduling. Our research stems from the practical goal of autonomous drone navigation and achieves superior autonomy and navigation performance by meticulously synergizing the navigation model and auxiliary model.

\textbf{Vision Language Model in drone navigation}
Recently, VLMs have garnered significant attention in the field of autonomous drone navigation due to their capability to jointly process visual and linguistic inputs, enabling high-level scene understanding and decision-making. Several studies \cite{han2025multimodalfusionvisionlanguagemodels,ma2025surveyvisionlanguageactionmodelsembodied,Enablingvison} explore the integration of VLMs into robotic systems, demonstrating their effectiveness in complex perception tasks such as semantic mapping, obstacle avoidance, and goal-oriented navigation. In particular, recent works \cite{chatFly, wang2025gscepromptframeworkenhanced} further investigate the application of vision language models in drone missions by constructing task reasoning frameworks and streaming inference architectures. In contrast to conventional vision-based pipelines that depend on manually crafted visual features, VLMs learn semantic representations in an end-to-end manner, which has been shown to enhance perceptual generalization and support more reliable navigation in heterogeneous environments. Building upon these advances, our research incorporates the state-of-the-art vision language model Qwen-VL-Max into the drone's perception and decision-making pipeline, integrated with edge computing technologies, significantly enhancing the capability for safe and efficient autonomous cruising.
\section{Background and Motivation}
\label{sec:background_motivation}
Our objective is to optimize the navigation task through collaborative utilization of edge, terminal, and cloud-based foundation models. Toward this end, we first introduce the navigation task(reflects \ref{subsec:autonomy}) and the foundation model task(reflects \ref{subsec:depth} and \ref{subsec:vlm}), both of which are inherently integrated into the edge-assisted autonomous navigation system. The design objectives of the system are formally characterized through specified quality-of-navigation (QoN) metrics(reflects \ref{subsec:qon}), which serve as key performance indicators for evaluating the effectiveness of the proposed framework.

\subsection{Autonomous Drone Navigation}
\label{subsec:autonomy}
Autonomous navigation methods function as an enabler for the safe deployment of drones on a wide range of real-world applications\cite{kouris2018learning}. The recent progress in DNN models has pushed DL-based navigation approaches to an unprecedented altitude. We focus on the self-sufficient navigation model for drones, wherein the generation of steering commands relies solely on computational resources available onboard the drone.

\begin{figure}[htbp]
    \centering
    \includegraphics[width=.9\linewidth]{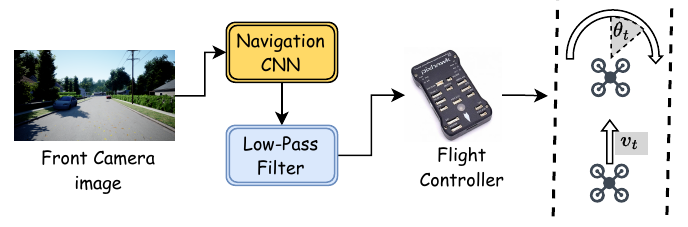}
    \caption{Autonomous navigation overview.}
    \label{fig:autonav_overview}
\end{figure}

The autonomous navigation task's goal is to navigate the drone through the unseen environment, achieving collision-free flight as long as possible. As illustrated in Figure \ref{fig:autonav_overview}, The workflow of the navigation task contains several components that collaborate to keep the drone on its intended path.
\begin{itemize}
    \item \textit{Navigation NN.} The DNN model takes the front-captured images as the input and exports a corresponding navigation decision, comprising two parts. One is the steering angle $\theta_t \in [-1,1]$, which is specified in the turning radian concerning the current orientation and will be used to direct the turning obliquity of aerofoils at the next moment. The other one is collision rate $p_t \in [0,1]$ which is used to generate the drone’s forward velocity\cite{dronet_ral2018}. 
    \item \textit{Low-Pass Filter.} The steering angle $\theta_t$ represented by the radian system will be low-pass filtered to smooth the control instructions. Commands will be applied over the next few decision intervals, to keep the drone flying in the intended direction.
\end{itemize}
Due to the limited onboard computational resources of unmanned aerial vehicles (UAVs), shallow convolutional neural networks (CNNs) are only capable of performing basic obstacle avoidance tasks and lack sufficient environmental perception capabilities. Consequently, relying solely on collision rate as the output for adjusting the UAV's movement speed is insufficient to ensure safe navigation. Such an approach may lead to collisions in the presence of sudden environmental changes. To address this limitation, it is necessary to incorporate additional foundation models to assist the UAV in handling complex and dynamic scenarios effectively.

\subsection{Depth Estimation Model Assisted Navigation}
\label{subsec:depth}
In mobile navigation systems, including UAVs, estimating accurate dense depth maps is a critical task, as depth estimation provides finer-grained environmental perception \cite{monodepth17}. Propelled by depth maps, flying drones can obtain a pixel-level understanding of surroundings to achieve successful navigation. However, devices that directly acquire depth maps are generally heavy, expensive, and energy-intensive, making it impractical to mount such equipment directly on drones\cite{Fonder2022Parallax}. Consequently, the distance between objects and the camera must be inferred using models rather than measured directly. We leverage the Monocular Depth Estimation to achieve powerful open-set segmentation, which generates the depth map with the image as the input.
\begin{figure}[htbp]
    \centering
    \includegraphics[width=.9\linewidth]{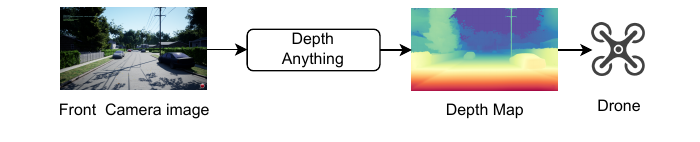}
    \caption{depth anything overview.}
    \label{fig:depth_overview}
\end{figure}
As illustrated in Figure \ref{fig:depth_overview}, The camera mounted on the drones consistently captures and publishes images when necessary. The computational units execute the depth model using the received image as input and generate the depth map. Based on the environmental depth information provided by the current depth image, the drone performs navigation corrections.

Notably, in drone-based data collection, the drone itself is equipped with onboard computational units capable of participating in data processing. However, due to the limited computational power of the onboard units, following the approach of several pioneering works\cite{xiao2022dnn,yan2023acmmm}, we leverage edge computing to alleviate the conflict between the growing complexity of DNN and the limited resources available on drones,  In this framework, the drone is responsible for transmitting captured images and decides whether to offload tasks to an edge server (e.g., a 5G MEC server or a WiFi-connected edge server) for input image collection and depth model computation based on the current network conditions.

\subsection{Vision Language Model For Navigation}
\label{subsec:vlm}
In mobile navigation systems, vision-language models (VLMs) enable semantic-level understanding of both visual inputs and user commands\cite{2019arXiv190802265L}, thereby generating appropriate flight instructions for unmanned aerial vehicles (UAVs), making them a promising solution for autonomous navigation. However, due to the typically large parameter sizes of VLMs, even deploying a relatively lightweight 7B-parameter model on edge servers poses considerable computational challenges. Moreover, the inference process often entails substantial latency, and the resulting outputs require additional time for parsing—limitations that hinder their direct application in real-time UAV navigation scenarios. Inspired by several pioneering works\cite{2020arXiv200514165B, 2022arXiv220108239T}, we adopt a cloud computing paradigm and employ an efficient invocation strategy to access VLMs. Furthermore, we utilize function-calling mechanisms to streamline the interpretation of inference results, thereby reducing processing overhead and improving responsiveness.


\subsection{Terminology}
\textbf{Quality of Navigation (QoN).} 
\label{subsec:qon}
Following the wisdom from \cite{chen2022adadrone}, we treat autonomous navigation as a service and extend the concept of Quality of Navigation (QoN) by introducing the depth map as the reference. We instantiate the SLO as a lower depth bound threshold $\varepsilon$ in the current steering angle, indicating the acceptable level of tolerance regarding navigation precision. Specifically, At any given time $t$, the current depth map can be noted as $\mathbf{X}_{t}$. The navigation model predicts the steering angle, denoted as $\theta_{t}^{pre}$, based on the previous input image captured at time $t-t_{delay}$, where the $t_{delay}$ represents decision delay. Then the steering angle is directly mapped to the depth value $\mathbf{X}_t(g(\theta_{t}^{pre}))$ in the direction derived by the angle-direction transformation $g(\cdot)$. The navigation service should satisfy:
\begin{equation}
    \mathbf{X}_t(g(\theta_{t}^{pre})) \geq \varepsilon. 
\end{equation}
Then the Quality of Navigation(QoN), denoted as $\mathcal{Q}_n$, is interpreted as how many times the navigation decision meets the SLO within a given time window $\tau$:
\begin{equation}
    \mathcal{Q}_n = \sum_{t=0}^{\tau}I(\mathbf{X}_t(g(\theta_{t}^{pre})) \geq \varepsilon)/\tau,
    \label{metric:qon}
\end{equation}
where $I(\cdot)$ denotes an indicator function that yields a value of 1 when the predicate evaluates to true.

\section{System Design}
\label{sec:codrone}

\begin{figure}[t]
  \centering
  \includegraphics[width=0.99\linewidth]{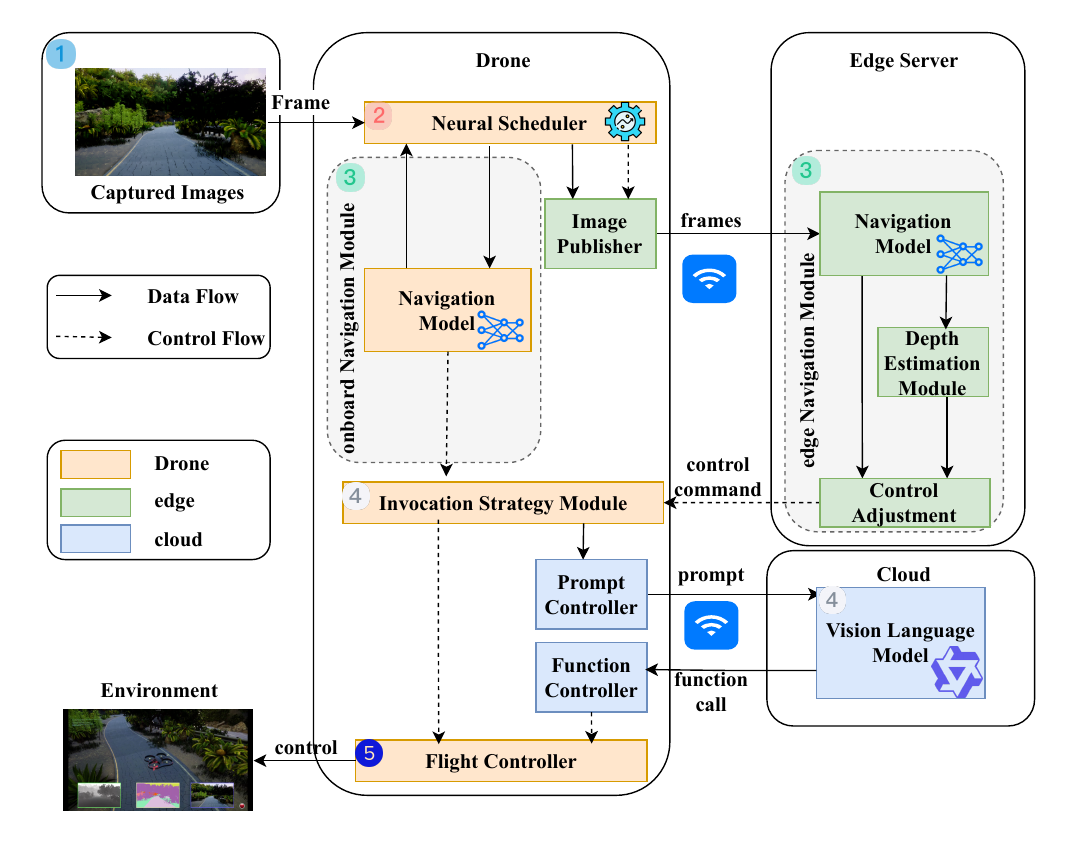}
  \caption{The overview of CoDrone in an end-edge-cloud collaborative framework.}
  \label{fig:codrone_overview}
\end{figure}

\subsection{System Overview}
CoDrone systematically leverages a hierarchical computing architecture (end-edge-cloud layers) to enable seamless, multi-granular control adjustments for UAV navigation, targeting both high-precision navigation and flight safety. Figure \ref{fig:codrone_overview} depicts the system architecture, where: (i) the \textbf{drone} (orange) executes real-time navigation control through its onboard module, which processes visual input to generate steering commands and manages task offloading via neural scheduling for efficient flight in dynamic environments; (ii) the \textbf{edge server} (green) serves as an acceleration platform, it processes imagery captured by UAVs in real time, selectively enhancing environmental perception through a depth estimation module and computing critical navigation parameters. By offloading computation-intensive tasks from the drone’s onboard system, the proposed approach ensures efficient deep neural network inference, thereby enabling autonomous flight even under stringent resource constraints; and (iii) the \textbf{cloud} (blue) operates semantic reasoning workflows via VLM module. Leveraging the strong image and language understanding capabilities of Vision-Language Models (VLMs), the cloud module ensures safe UAV operation in extreme scenarios by directly invoking low-level flight control functions. In the following, we outline the system's operational procedure in detail.

During autonomous navigation, the drone \ding{192} interacts with the physical environment, continuously capturing images via its front-facing camera to inform subsequent navigation decisions. The \textbf{Neural Scheduler Module} deployed on the UAV \ding{193}, generates task scheduling directives based on historical flight data and real-time network bandwidth measurements. These directives determine the execution strategy, including: processing location (onboard or edge offloading), visual data compression ratio during offloading, and activation status of the depth estimation model for enhanced navigation assistance. Once the execution strategy is determined, the \textbf{Navigation Module} inducing onboard navigation module and edge navigation module, \ding{194} processes incoming visual data correspondingly. Following the Neural Scheduler’s instructions, it infers critical navigation parameters (e.g., steering angles and collision probabilities), enabling real-time flight control and maneuver decision-making. Due to the navigation module's lack of semantic perception capabilities, the \textbf{Invocation Strategy Module} \ding{195} evaluates the environmental complexity by analyzing the navigation model’s outputs and historical scheduling decisions, determining whether to engage the cloud-based Vision-Language Model. When activated, the VLM integrates the transmitted visual data with preprocessed function prompts and semantic instructions to infer the optimal action and directly invokes the corresponding functions to assist the UAV in navigating through complex or critical scenarios. Finally, all inference results are aggregated by the \textbf{Flight Controller} \ding{196}, which executes flight commands based on the received information to control the UAV’s movement.


\subsection{Edge-assisted Drone Navigation Design}
Although edge servers generally offer superior computational resources (e.g., GPUs) and faster inference speeds compared to UAV onboard systems, the presence of network latency and stringent real-time constraints makes it infeasible to offload all tasks to the edge server. To address this challenge, we deploy heterogeneous navigation modules on both the UAV and the edge server. Specifically, the onboard navigation module consists solely of a lightweight deep neural network (DNN), while the edge-based counterpart incorporates both the lightweight DNN and an additional depth estimation model.
Inspired by common practices in edge-assisted frameworks \cite{chen2022adadrone, Uav-assisted}, our system selectively offloads computation-intensive navigation tasks to the edge server via Wi-Fi or 5G under favorable bandwidth conditions, thereby optimizing runtime efficiency. Unlike conventional edge-assisted UAV navigation architectures, our approach enhances environmental perception capabilities by integrating a depth estimation module on the edge server, which is activated when necessary to provide fine-grained spatial understanding. This design contributes to more reliable decision-making in complex or dynamically changing environments by supplying additional spatial cues when required.
The proposed system architecture comprises three core components:
\textbf{Onboard Navigation Module} directly processes monocular camera input to infer steering angles and collision probabilities; 
\textbf{Edge Navigation Module} extends the Onboard Navigation Module with an additional Depth Estimation Model, enabling enhanced environmental perception. Upon instruction from the Neural Scheduler Module, the Edge Navigation Module selectively converts image data into an occupancy grid map to accurate real-time navigation parameter estimation and enhances obstacle avoidance capabilities;
\textbf{Neural Scheduler Module} coordinates task scheduling between the edge server and UAV.

\subsubsection{Onboard Navigation Module}
Due to the strict real-time demands of autonomous navigation tasks, the system must process visual input efficiently and generate accurate control signals within tight time constraints. However, the limited payload and battery capacity of drones make it impractical to deploy high-power computing hardware onboard. Consequently, the Onboard Navigation Module adopts a lightweight DNN architecture, selected to reduce computational overhead while preserving a reasonable level of navigation performance under real-time constraints. To further reduce the onboard computational burden while balancing navigation accuracy and efficiency, we use an optimized input data pipeline for DNN-based image inference.

RGB images from monocular cameras generally contain three channels, where the chrominance channels primarily convey color information of the surrounding environment, while contributing relatively little to the representation of the intrinsic features of objects within the scene. However, the requirement to display color necessitates the inclusion of two additional channels, significantly increasing the portion of the storage space. Experimental results show that, across various resolutions (ranging from $112^2$ to $448^2$ pixels), the memory footprint of RGB images increases by 50\% compared to grayscale images at a resolution of $224^2$ pixels (increasing from 48KB to 120KB), with this disparity becoming more pronounced as pixel counts increase. For onboard inference, grayscale conversion reduces the input channel dimension by $66\%$, directly decreasing matrix computation complexity by approximately the same proportion. 
This substantial reduction is particularly critical for resource-constrained onboard processing. In edge-offloading scenarios, grayscale images also reduce transmission latency due to their smaller size. Consequently, we adopt grayscale images for training and inference with the onboard DNN navigation model to further minimize the computational burden on the drone.


\subsubsection{Edge Navigation Module}

Conventional RGB images and grayscale pictures merely provide visual information without indicating the precise distance of objects from the observer. In contrast, depth maps offer three-dimensional information about the surrounding environment, enabling UAVs to accurately perceive the distances between themselves and obstacles\cite{MonocularDepthEstimation, FastMonocular}. Given the substantially greater computational capacity available on edge servers relative to UAV onboard platforms, these servers can support the execution of deeper and more computationally demanding DNN models.
Accordingly, the Edge Navigation Module extends the functionality of the Onboard Navigation Module by integrating an additional depth estimation model. 
When enhanced environmental perception is required to handle potential obstacles, the Neural Scheduler Module dynamically activates the Depth Estimation Module. This module first infers a depth map from the input visual image, which is then transformed into a one-dimensional occupancy grid map using the DEGAGE algorithm. Based on this occupancy grid map, the outputs of the navigation model are subsequently refined to improve path planning and obstacle avoidance performance.

This process enables precise localization of obstacles in terms of both spatial position and distance, thereby enhancing the system’s fine-grained environmental awareness. By integrating the Depth Estimation Module into the overall architecture, the perceptual limitations of shallow DNN models under resource-constrained conditions are effectively mitigated, leading to improved navigation robustness and situational awareness.

To derive the current depth information, the drone first acquires an image $\mathbf{X}^{rgb}_{t}$ from the RGB camera in the $t$ moment, $\mathbf{X}^{rgb}_{t} \in \mathbb{R}^{h \times w \times c}$, where $h$, $w$, and $c$ refer to the height, width, and number of channels of the image, respectively. Then the image is transmitted to the edge server, and the edge server sends the image into the depth estimation model for inference to obtain the depth estimation, defined as:
\begin{align}
    \mathbf{X}^{Depth} &= H_{\text{Depth}}(\mathbf{X}^{rgb}_{t}, \Theta),
\end{align}
where $H_{Depth}(\cdot)$ is the depth estimation model and $\Theta$ represents the parameters of the model. $\mathbf{X}^{Depth} \in \mathbb{R}^{h \times w}$ denotes the generated masks. it can't be used for decision-making directly, as the distances represented by each pixel vary significantly.to better assess the distance between the current obstacle and the drone's position, we apply a targeted rounding operation to the obstacle distances for subsequent judgment.
\begin{align}
    \mathbf{D} &= \lfloor \mathbf{X}^{\text{depth}} \rfloor \in \mathbb{Z}^{h \times w} \label{eq:quantization},
\end{align}
where $\lfloor\cdot\rfloor$ represents the Flooring operation for integer conversion. Specifically, in the depth map $\mathbf{D} \in \mathbb{R}^{h \times w}$, each pixel value encodes the relative distance between the corresponding scene point and the camera. Leveraging this distance information, we can directly infer the spatial configuration of obstacles relative to the camera in the operational environment. Subsequently, this positional relationship is employed to validate the correctness of the current navigation deviation angle. By analyzing the relative depth relationships among different pixels, we can classify the map into three distinct categories to ensure safe navigation:

\begin{itemize}
    \item[1.] \textit{Drivable region}: A relatively empty road surface without collisions, an unobstructed environment, etc. 
    \item[2.] \textit{Low-collision-risk region}: An obstacle that needs to be avoided, but is far away. 
    \item[3.] \textit{High-collision-risk region}: An obstacle that is highly likely to collide. 
\end{itemize}

Leveraging the widely adopted fixed-altitude flight mode in autonomous UAV navigation systems—where vertical depth information contributes minimally to horizontal obstacle avoidance—we propose a novel occupancy map generation algorithm (see Algorithm \ref{algo:desga}). The core idea of this method is to perform vertical discretization on depth images, transforming raw depth data into a physical context that aligns directly with the UAV’s horizontal flight trajectory. Through one-dimensional discrete sampling, the algorithm distinguishes ahead regions as either flyable areas or High-collision-risk regions, thereby assisting the UAV in perceiving potential hazards. Compared to traditional two-dimensional segmentation maps, this one-dimensional representation significantly reduces data overhead while enabling faster and more accurate safety assessments, thus enhancing overall flight safety and efficiency.

\begin{algorithm}
\label{algo:desga}
\caption{Depth Estimation Map Generation Algorithm(DEGAGE)}\label{DEMGA_algorithm}
\KwData{Depth Estimation map $\mathbf{D}$}
\KwResult{occupancy map $\mathbf{o} \in [0,1]^{w}$}
$\mathbf{D}^{center} =  f_{center}(D) $\;
$\mathbf{w} \leftarrow \mathbf{W}$ in $D^{center}$
$\mathbf{o} \leftarrow \mathbf{1} \in [0,1]^{w}$\;
$i \leftarrow 0$\; 
\For{$i \leftarrow 0$ \KwTo $W$ in $D^{center}$ }{
    \For{$j \leftarrow 0$ \KwTo $H$ in $D^{center}$}{
        \If{$\mathbf{D}^{center}[j,i] \text{ is Drivable region}$}{
            $\mathbf{O}[i] \gets 0$; 
        }
        \Else{
            $\mathbf{o}[i] = f(\Delta_{area}(i))$\;
        }
    $j \leftarrow j + 1$;
    }
    $i \leftarrow i + 1$;
}
\end{algorithm}

Lines 4-15 that if certain regions displayed in the image are obstacles, i.e., not navigable areas, the area change of the region represented by the $i$-th pixel point is recorded between consecutive frames, denoted as $\Delta_{area}(i)$, $f(\cdot)$is a monotonically increasing mapping function, significant changes in the area transformation indicate a heightened risk of collision. This binary flagging mechanism indicates that continued steering or forward motion in that direction carries a significant risk of collision, as the depth map has been preprocessed with vertical cropping based on the UAV’s flight altitude to isolate critical spatial constraints.

Upon generation of the occupancy map, depth information regarding obstacles in the forward environment becomes available. This information facilitates the refinement of control commands predicted by the navigation model, ultimately ensuring collision-free flight. A greedy algorithm (Algo.\ref{algo:caa}) is adopted to achieve this control command refinement process.

\begin{algorithm}
\label{algo:caa}
\caption{Control Adjustment Algorithm(CAA)}\label{CAA_algorithm}
\KwData{current period $t$, yaw angle $s_t \in [-1,1]$, collision probability $p_t \in [0,1]$, occupancy map $\mathbf{o} \in [0,1]^{w}$, width of drone $\hat{w}$, maximal flying speed $V_{max}$, high collision probability threshold $T_{Hrisk}$, low collision probability threshold $T_{Lrisk}$}
\KwResult{adjusted yaw angle $\theta_{t}$, adjusted speed $v^{'}_{t}$}
$\alpha \leftarrow 0.5$\;
$\beta \leftarrow 0.5$\;

$v^{'}_{t} = \alpha v^{'}_{t-1} + (1-\alpha)(1-p) V_{max}$ \;
$\theta_{t} = \beta\theta_{t-1} + (1 -\beta) \frac{\pi}{2} s^{'}_{t}$ \;

\If {$ \mathbf{o}[\frac{(s^{'}_{t}+1)}{2} w - \frac{\hat{w}}{2}: \frac{(s^{'}_{t}+1)}{2} w + \frac{\hat{w}}{2}]  >= T_{lrisk}$}{
\If{$\mathbf{o}[\frac{(s^{'}_{t}+1)}{2} w - \frac{\hat{w}}{2}: \frac{(s^{'}_{t}+1)}{2} w + \frac{\hat{w}}{2}] > T_{hrisk}$}{$v^{'}_{t} \leftarrow 0$\;}
$v^{'}_{t} \leftarrow \alpha v^{'}_{t}$\;
}
\end{algorithm}

\begin{figure}[htbp]
    \centering
    \includegraphics[width=1\linewidth]{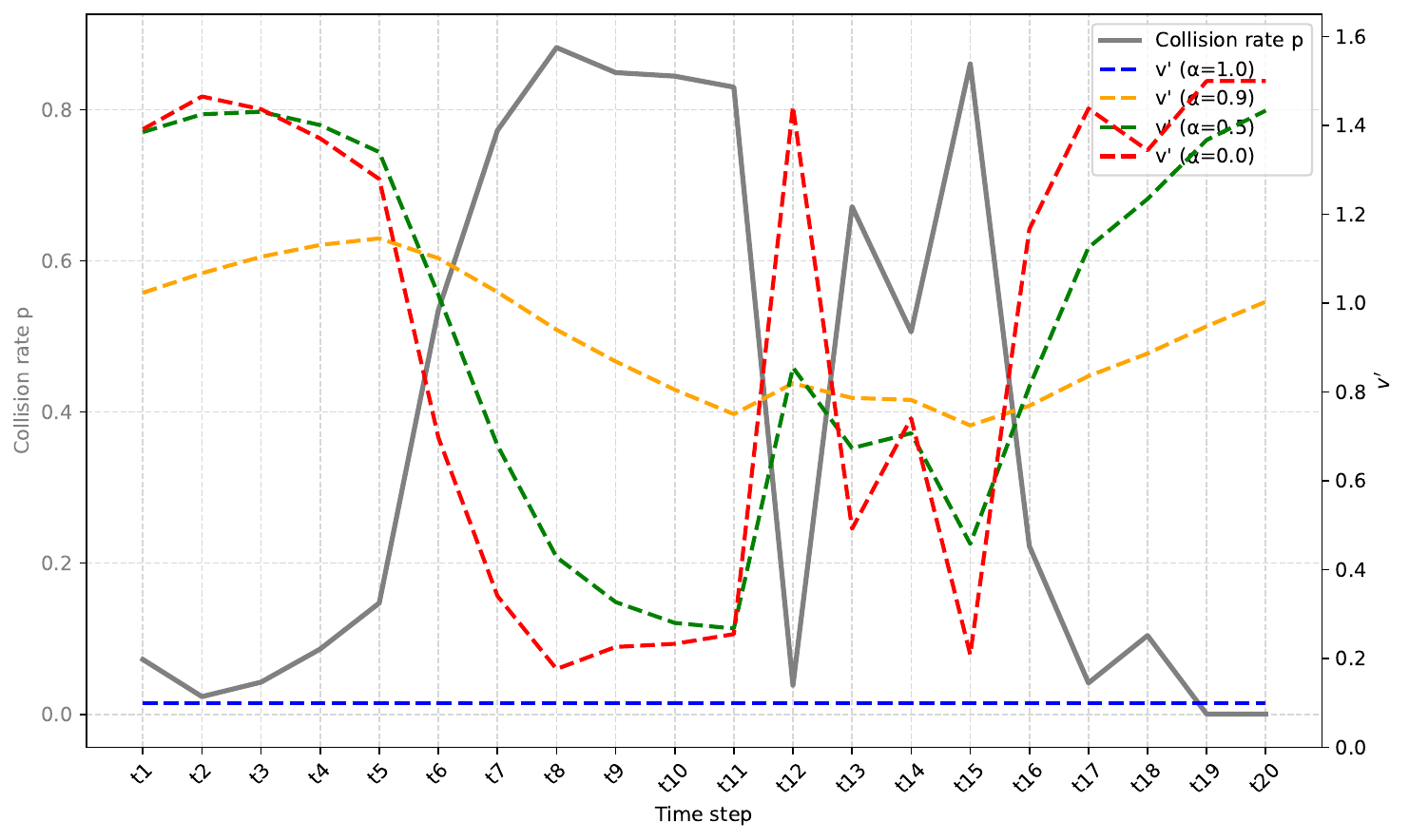}
    \caption{Plot of Velocity v' Variation with Time Step and Collision Rate p for Different Values of Parameter $\alpha$}
    \label{fig:Parameter}
\end{figure}

Lines 3-4 describe the implementation of a low-pass filter. In the low-pass filter, $\alpha$ serves as a smoothing coefficient in the velocity update formula:
\begin{equation}
    v_t' = \alpha \cdot v_{t-1}' + (1-\alpha) \cdot (1-p) \cdot V_{\text{max}},
\end{equation}
where $v_t'$ and $v_{t-1}'$ represent the current and previous filtered velocities, respectively; $p$ denotes the measured collision rate; and $V_{\max}$ is the maximum allowable velocity magnitude. The coefficient $\alpha \in [0,1]$ serves as a smoothing factor that determines the relative weighting between the historical velocity term $\alpha v_{t-1}'$ and the collision-rate-driven correction term $(1-\alpha)(1-p)V_{\max}$.
A larger value of $\alpha$ emphasizes the historical velocity, thereby enhancing smoothing and reducing high-frequency fluctuations. However, excessive smoothing diminishes the model's responsiveness to collision-rate variations. Conversely, a small value of $\alpha$ increases the influence of the most recent collision-rate measurement, improving responsiveness but potentially introducing instability or noise into the velocity trajectory.
To determine an appropriate value for $\alpha$, we conducted a parameter-sensitivity analysis under the conditions of an initial velocity magnitude of $0.1$ and $V_{\max}=1$. As illustrated in Fig.~\ref{fig:Parameter}, different values of $\alpha$ produce distinct velocity behaviors with respect to the collision rate. Specifically,
\begin{itemize}
    \item $\alpha=0$ results in excessive velocity oscillations due to complete reliance on instantaneous collision-rate measurements, failing to meet the stability requirement;
    \item $\alpha=0.9$ provides strong smoothing but insufficient sensitivity to collision-rate variations, preventing effective adaptation to environmental changes;
    \item $\alpha=0.5$ achieves a balanced trade-off between responsiveness and stability, reacting to collision-rate changes while maintaining a smooth velocity profile.
\end{itemize}

Based on these observations, $\alpha=0.5$ is selected as the default configuration. The same selection rationale is also adopted for the parameter $\beta$ in the corresponding update process.

Lines 5–10 elaborate on the command optimization process for occupancy map integration: the system evaluates whether the relative depth of obstacles within the field of view of the current steering angle exceeds a predefined threshold. If the area is identified as a high-collision-risk region, it indicates an imminent collision risk for the UAV, in which case the UAV’s speed must be promptly reduced to zero to ensure its operational safety. Whereas if the area is classified as a low-collision-risk region, the UAV is deemed to have a certain probability of collision; thus, its speed should first be reduced to half of its original value to prevent subsequent collisions caused by excessive speed.

When there is a collision risk, the risk recognition accuracy of using a depth estimation model to calculate occupancy grids based on depth is $R_{\text{depth}}$, and the risk recognition accuracy inferred solely from RGB images is $(R_{\text{rgb}})$. Since depth estimation images judge collisions based on relative depth, $R_{\text{depth}} \geq R_{\text{rgb}}$. 
Since the collision probability is positively correlated with both the risk recognition error and the speed. According to Algorithm CAA, the expected collision probability of the depth-based method is:
\begin{equation}
\mathbb{E}[P_D] \propto \alpha v^{'}(1- R_{\text{depth}}).
\end{equation}

The expected collision probability of the RGB-based method is:
\begin{equation}
\mathbb{E}[P_R] \propto (\alpha v^{'} + (1-\alpha)(1-p) V_{max} ) \cdot (1 - R_{\text{rgb}}).
\end{equation}

Since $ R_{\text{depth}} > R_{\text{rgb}} $ and $ \mathbb \alpha v^{'} + (1-\alpha)(1-p) V_{max} > \alpha v^{'} $, it follows that $ \mathbb{E}[P_D] < \mathbb{E}[P_R] $.Therefore, using depth estimation as an auxiliary when appropriate can reduce the collision risk of UAVs.

\subsubsection{Neural Scheduler Module}
\begin{figure}[htbp]
    \centering
    \includegraphics[width=1\linewidth]{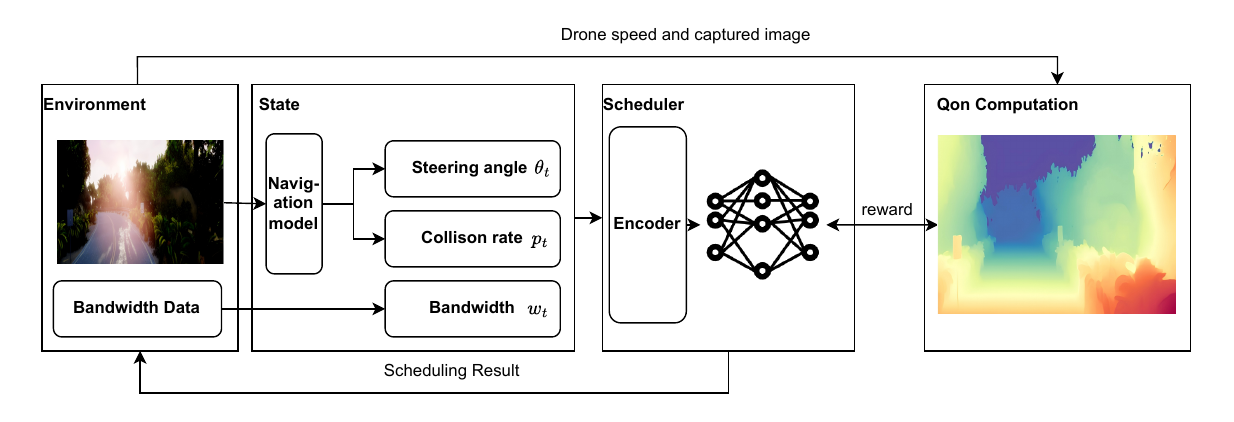}
    \caption{CoDrone Neural Scheduler}
    \label{fig:drl}
\end{figure}

The Neural Scheduler Module orchestrates intelligent computation offloading for navigation-related tasks in UAV systems. During the real-time flight, it dynamically determines task execution locations based on multiple real-time factors, including end-edge network bandwidth, current collision probability, and steering angle dynamics. For instance, when bandwidth is limited and collision risk is high, edge offloading, while computationally advantageous, may introduce prohibitive transmission latency, potentially compromising obstacle avoidance. To address these challenges, To mitigate these challenges, this module continuously assesses system states and optimizes offloading strategy, which includes visual data compression ratios for transmission and context-aware activation of the depth estimation model to augment environmental perception. Notably, the module resolves coordination challenges across heterogeneous computing layers (device-edge-cloud) and between primary navigation and auxiliary foundation models. As a core enabler of the end-edge-cloud framework, it seamlessly integrates edge resources with onboard systems to enable efficient orchestration of inference tasks. This design ensures bounded latency and accurate invocation of foundation model-based navigation assistance, even under network instability and complex environmental variations.

 As illustrated in Figure \ref{fig:drl}, the neural scheduler captures the collision rate, steering angle information, and bandwidth information output by the previous model, and generates the scheduling scheme for the next inference. Subsequently, iterative optimization is performed based on the results of the next inference model and the depth estimation map corresponding to the front-facing camera image captured after obtaining the aforementioned results.given the uncertainty in drone flight states (steering angle $(\theta_t)$, collision probability $(p_t)$), the average collision rate over the last 5 steps($ap_t$) and network fluctuations (bandwidth $(w_t)$), CoDrone’s scheduling involves a large search space, further complicated by the irregular and non-smooth optimization objective,  precluding the possible application of the heuristic mathematical methodology. Conventional mathematical optimization algorithms prove inadequate for this problem, as they inherently rely on the availability of accurate, static system models and require objective functions that are both differentiable and structurally well-behaved. Consequently, they cannot effectively accommodate non-smooth or discontinuous optimization objectives.To this end, we propose a Deep Reinforcement Learning (DRL)-based scheduling algorithm tailored for CoDrone. The scheduling problem can be formally formulated as a Markov Decision Process (MDP), with detailed definitions of state, action, and reward functions as follows:

\textbf{State}: The observable environmental information at time $t$, including the output of the navigation model (steering angle $\theta_t$ and collision rate $p_t$), the average collision rate $ap_t$ over the last 5 steps, bandwidth $w_t$ between the drone and the edge. The state of the neural scheduler consists of $\langle\theta_t, p_t, ap_t, w_t\rangle$.

\textbf{Action}: The specific configuration $\langle l, o, c\rangle$ characterizes the runtime configuration in the system. $l$ represents which inference execution location should be selected at the current moment, $o$ represents whether to use the depth estimation model at the current moment, and $c$ is the compression rate set when the drone pushes the image. We discretize the action space for simplicity, where $l \in \{0,1\}$ (0 for on board and 1 for edge server), $o \in \{0, 1\}$(0 for need depth estimation model assist and 1 for only need navigation model), and $c \in \{95\%, 80\%, 60\%\}$.

\textbf{Reward}: The scheduling agent's goal is to maximize the cumulative expected reward QoN defined by Eq.\ref{subsec:qon}. Given a fixed model inference accuracy, QoN is correlated with inference latency and total flight time; this effectively renders the optimization of QoN equivalent to balancing end-to-end inference latency and the utilization of depth estimation-aided models. Notably, in practical deployment scenarios, explicit computation of QoN is unnecessary; instead, the scheduler can directly map the current system state to an appropriate scheduling action through learned policy or rule-based mechanisms.

\begin{table}[t]
\centering
\renewcommand{\arraystretch}{1.5}
\caption{Comparison of Input Tensor Sizes and Computational Costs}
\label{tab:flops_compare}
\begin{tabular}{lcc}
\hline
\textbf{Model} & \textbf{Input Tensor Size} & \textbf{FLOPs} \\
\hline
Navigation Model & $(1,\,200,\,200,\,1)$ & $4.3\times10^{7}$ \\
Scheduler Model  & $(1,\,4)$             & $1.0\times10^{4}$ \\
\hline
\end{tabular}
\end{table}

We employ the off-the-shelf A2C algorithm\cite{mnih2013playing} to train our scheduler. As illustrated in Table~\ref{tab:flops_compare}, the computational overhead introduced by the scheduler is only a tiny fraction of that required by the navigation model. This large discrepancy mainly results from the substantial difference in their input sizes: the navigation network processes a high-resolution tensor of size, whereas the RL scheduler receives only a compact state vector of dimension. Consequently, the integration of the RL module imposes minimal additional computational burden.. To accelerate the training process, we customize a simulation environment based on the gym library\cite{brockman2016openai}. Specifically, we use the Mid-Air\cite{fonder2019mid} dataset and the HSDPA\cite{riiser2013commute} dataset to simulate the drone flight environment and fluctuations of edge networking conditions during flight. We also profile the inference latency using the Jetson Nano as an onboard computational device and the workstation with GPU as the edge server.

\subsubsection{Discussion}
The edge-assisted UAV navigation system leverages the Neural Scheduler Module to enable intelligent task offloading, image compression, and invocation of auxiliary foundation models, thereby achieving seamless integration between onboard computation and edge server capabilities. More importantly, the incorporation of the depth estimation module enhances the system’s environmental perception with fine-grained spatial awareness beyond traditional end-edge collaboration frameworks. This module complements the navigation model by providing detailed obstacle localization, thus enabling more accurate and safer navigation decisions.

While end-edge collaborative frameworks demonstrate improved decision-making capabilities compared to monolithic architectures, their efficacy remains constrained in safety-critical scenarios. Specifically, when UAVs encounter unexpected environmental obstacles (e.g., inadvertent flight into restricted green belts), existing systems exhibit fundamental limitations: (1) onboard navigation models typically output binary collision probability estimates without contextual reasoning, and (2) depth estimation modules merely provide low-level occupancy information without semantic interpretation. This capability gap often leads to navigation deadlock, where the UAV remains trapped in unresolvable situations despite possessing partial environmental awareness. The root cause can be attributed to the inherent limitation of light-weight DNNs in achieving semantic-level environmental understanding a critical shortcoming that persists across most current UAV autonomy solutions.

Vision language models (VLMs), empowered by their strong visual-semantic reasoning capabilities, offer a promising solution to this challenge\cite{shinde2025surveyefficientvisionlanguagemodels}. By leveraging their contextual visual–semantic reasoning capabilities, these models can improve the system’s capacity to interpret complex environments and support more reliable navigation decision-making. However, edge servers typically suffer from limited computational resources, making it infeasible to deploy large-scale VLMs directly at the edge. For instance, current mainstream VLMs, such as LLaVA-7B or BLIP-2, typically require at least 20GB of GPU memory to perform inference \cite{llava}, which far exceeds the typical GPU configurations available on most edge devices (usually ranging from 8GB to 16GB of memory). When opting for smaller, lightweight variants (e.g., distilled models with 1–3 billion parameters) to meet resource constraints, the performance on vision-language cross-modal understanding tasks degrades significantly. Therefore, direct deployment of large VLMs on edge servers is generally impractical. Instead, we adopt a cloud-based deployment strategy, leveraging a pre-deployed VLM in the cloud to fulfill the semantic reasoning requirements of our system.

\subsection{Cloud-assisted Drone Navigation Design}
While VLMs are known for their strong semantic understanding capabilities and have been independently applied in various UAV scenarios\cite{chatFly, saxena2025uavvlnendtoendvisionlanguage}, their direct integration into real-time autonomous navigation systems remains challenging due to high computational demands and long inference latencies. Unlike conventional approaches that treat VLMs as standalone decision-making modules, this work introduces a novel paradigm in which the VLM serves as a high-level assistant within an end-cloud collaborative framework. This design gives rise to a hybrid autonomous navigation system, where the edge-terminal architecture constitutes the primary processing layer, and the cloud-hosted VLM provides contextual support when necessary.
This cloud-assisted approach effectively addresses the limitations of edge-terminal devices, which often lack sufficient computational resources to deploy full-featured vision-language models. Furthermore, relying solely on lightweight DNNs or domain-specific models has been shown to be insufficient for accurately interpreting extreme environments and generating appropriate responses. By strategically invoking the VLM at critical decision points, the proposed cloud-based architecture delivers advanced perception and reasoning capabilities exactly when they are most needed, thereby significantly enhancing the robustness and adaptability of the autonomous navigation system.

The proposed cloud-level intervention mechanism comprises two core components: the Invocation Strategy Module (ISM) and the Vision-Language Model Module (VLM Module). The ISM dynamically determines whether and when to engage the VLM based on real-time environmental perception and system status. The VLM Module, on the other hand, enables high-level semantic interpretation of visual inputs by integrating UAV-specific flight functionalities with natural language prompts. This integration allows the system to generate context-aware control commands tailored to complex or unforeseen operational scenarios.

\subsubsection{Invocation Strategy Module}
The inference and utilization of visual language models are notably time-consuming. Additionally, directly invoking APIs for these models on the cloud is relatively costly at this stage. Therefore, during the process of drone cruising, to minimize expenses, it is impractical to frequently call upon visual language models for auxiliary support. To address the issue of invocation granularity, we integrate it with the aforementioned depth estimation results. It is posited that the depth estimation maps generated by the depth estimation model accurately reflect the obstacles ahead in the drone's environment. Given this, the DRL-based Neural Scheduler Module above learns the optimal moments requiring depth estimation map assistance based on collision rates and steering angles. These moments also represent the times when drones are likely to encounter collisions. Hence, under two circumstances, we may select the visual language models for assistance:

\begin{enumerate}
    \item A sudden increase in the predicted collision probability, accompanied by a significant deviation in the steering angle, may indicate that the drone has encountered an unforeseen obstacle. For instance, when a UAV is navigating through a previously clear corridor and suddenly detects a high collision risk along with abrupt leftward or rightward trajectory adjustments, it suggests the presence of an unexpected environmental change, such as the appearance of a partition or a moving object.
    \item If the reinforcement learning-based scheduling network consecutively selects the depth estimation model twice for navigation assistance, and the corresponding occupancy maps consistently indicate high-collision-risk regions directly ahead of the drone, it suggests that the UAV may be encountering a densely cluttered environment. In such scenarios, purely relying on geometric information becomes insufficient for safe navigation. Therefore, acquiring a semantic-level perception of the surroundings becomes essential to enable more informed and context-aware flight decisions.
\end{enumerate}

These two scenarios indicate a high probability of imminent collision and suggest a complex environmental condition. Consequently, the system suspends the drone and transmits current image information to the visual language model for further analysis and decision-making support.

\subsubsection{Vision Language Model Module}
Small, lightweight DNN models are widely adopted in UAV navigation systems to estimate critical parameters such as collision probabilities and optimal steering angles based on the drone’s immediate visual input. These models are often complemented by depth estimation modules that enhance obstacle detection and spatial awareness. However, a key limitation of such lightweight architectures is their inability to robustly detect deviations from a predefined flight trajectory or to adaptively respond to high-level navigational goals under dynamic environmental conditions. In these cases, it becomes necessary to leverage the advanced visual perception capabilities of VLMs. By invoking appropriate tools, the VLM can directly assist in controlling the drone to handle complex environmental conditions, thereby ensuring safe and efficient flight operations.
\paragraph{Function Call Enables Drone Cruising}
Recent advances in VLMs offer a promising solution to the above-mentioned challenges by enabling a rich semantic understanding of both visual inputs and natural language instructions. Unlike conventional DNN-based approaches, VLMs can interpret task-oriented textual commands such as “fly to the red door ” and contextualize them with real-time onboard imagery. This capability enables the system to formulate high-level navigational directives, encompassing strategic route refinement and obstacle avoidance actions informed by the surrounding context. For example, given an image captured by the drone’s front-facing camera and the instruction “move to the center of the roadway”. The VLM can analyze the scene, identify the junction structure, and determine whether the drone has reached the appropriate decision point. Based on this interpretation, it can then recommend a suitable maneuver.

Despite their capabilities, directly translating VLM outputs into actionable control signals remains challenging. Conventional methods rely on rule-based parsers or normalization functions to process linguistic outputs, introducing computational overhead and latency-critical drawbacks for resource-constrained aerial platforms. To address this challenge, we use a function call mechanism that tightly integrates the reasoning capabilities of the VLM with low-level drone control primitives. Unlike generic VLMs, we customize a UAV-specific vision-language module tailored for autonomous navigation tasks. Specifically, we define a set of fundamental flight commands, such as move\_forward, turn\_left, and get\_current\_position, and encapsulate them as callable functions within the system architecture. These functions constitute executable control primitives that the VLM can selectively invoke in accordance with its semantic interpretation of both the operational environment and the mission directives.
\begin{figure}[htbp]
    \centering
    \includegraphics[width=.99\linewidth]{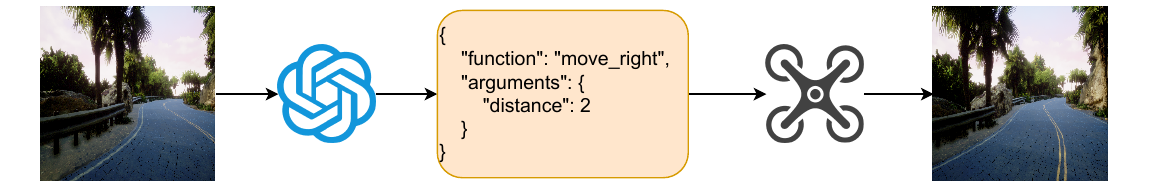}
    \caption{VLM Function Call Overview.}
    \label{fig:function call}
\end{figure}

The UAV-specific Vision-Language Module integrates multiple input modalities—including mission instructions, available flight functions, real-time positional data, and front-camera imagery to a unified toolchain. By jointly processing these heterogeneous inputs, the VLM generates not only a natural language explanation of its reasoning process but also a structured command in the form of a function name paired with its corresponding parameters(see Fig.~\ref{fig:function call}).
This direct generation of executable commands eliminates the need for additional post-processing or linguistic parsing stages, thereby significantly reducing execution latency and improving the overall responsiveness of the autonomous navigation system.

\paragraph{Prompt Design}

\begin{figure}[htbp]
    \centering
    \includegraphics[width=0.99\linewidth]{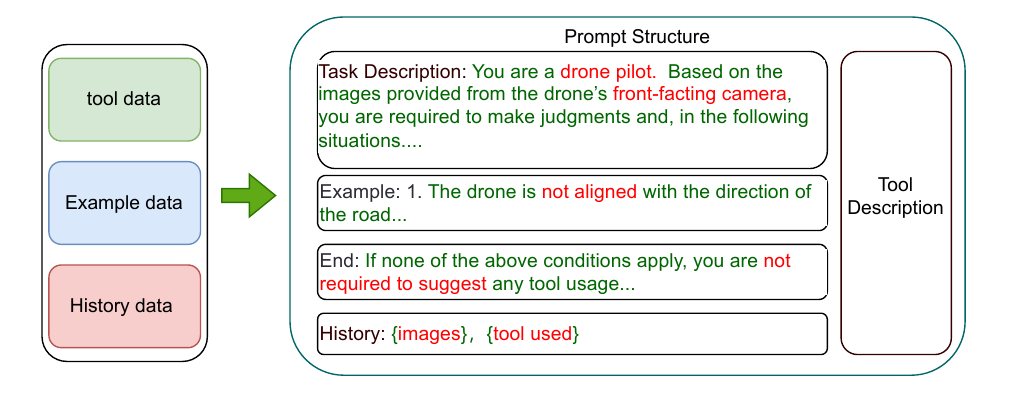}
    \caption{Prompt Structure}
    \label{fig:vlm-prompt}
\end{figure}
VLM is typically activated when a drone deviates from its intended trajectory or encounters an unmanageable collision risk through its conventional navigation system. Given the constraints of onboard computational resources and the prohibitive cost associated with fine-tuning large models for every specific flight scenario, the customization of pre-trained models is impractical. Consequently, the design of an effective prompt becomes paramount.

In the CoDrone system, prompt design comprises five essential components, as illustrated in Figure~\ref{fig:vlm-prompt}:

\begin{itemize}
    \item \textbf{Tool Description}: This component rigorously defines the available functions for drone control.
    \item \textbf{Task Description}: This explicitly mandates the VLM to operate as a drone pilot, aiming to re-establish the drone's correct flight path.
    \item \textbf{Example}: This provides concrete scenarios necessitating VLM intervention, thereby delineating the conditions under which the model should assume control.
    \item \textbf{End}: This specifies that no tools should be utilized if the current situation does not align with any of the defined examples.
    \item \textbf{History}: This incorporates recent visual data and prior tool invocation records to support sequential reasoning and improve decision consistency.
\end{itemize}

The inclusion of the history module is particularly crucial in complex and unfamiliar environments, where a single VLM function call is often insufficient for successful task completion. Instead, iterative reasoning and action execution guided by real-time visual feedback and previous outcomes are typically required. Furthermore, this module helps prevent repetitive or contradictory commands that could lead to oscillatory behavior or even deadlock. For example, without access to historical context, a VLM might issue a "turn left 90°" command based on the current visual input, and after execution, issue a conflicting "turn right 90°" command based on the new view, potentially causing an infinite loop. By incorporating past interactions, CoDrone ensures more coherent and consistent decision-making over time.

The proposed history module is activated whenever the VLM is engaged. It records both the front-facing camera images and the corresponding function calls made by the VLM. Upon completion of each function execution, the system incorporates both the prior visual input and the updated post-action imagery into the next prompt provided to the VLM. This mechanism constructs a complete contextual history for tool usage, enabling more coherent sequential decision-making. The inference process terminates only when the VLM stops issuing any further function calls, indicating that the assigned task has been accomplished.

\begin{figure}[t]
    \centering
    \includegraphics[width=.99\linewidth]{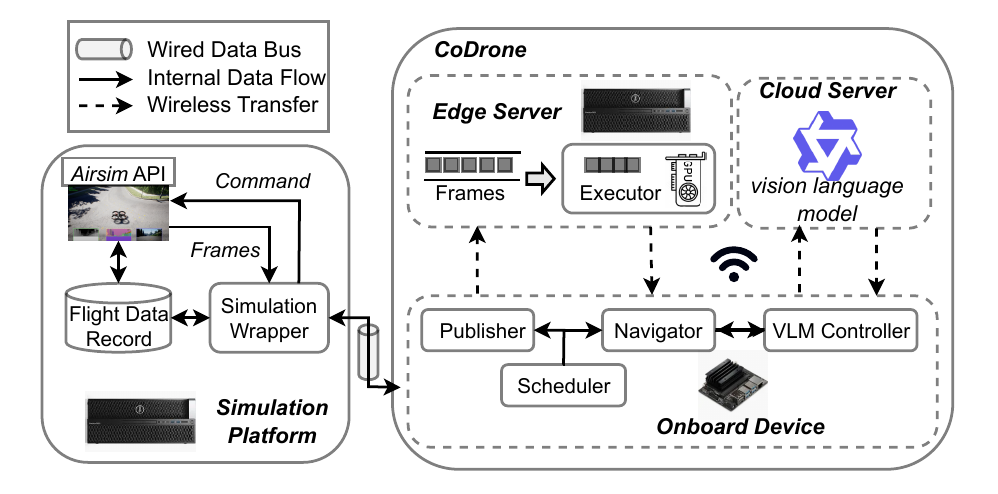}
    \caption{Implementation with Simulation Integration.}
    \label{fig:codrone_implementation_overview}
\end{figure}

\section{Experimental Implementation}
\label{sec:implementation}
\begin{figure}
    \centering
    \begin{minipage}{0.32\linewidth}
        \centerline{\includegraphics[width=\textwidth]{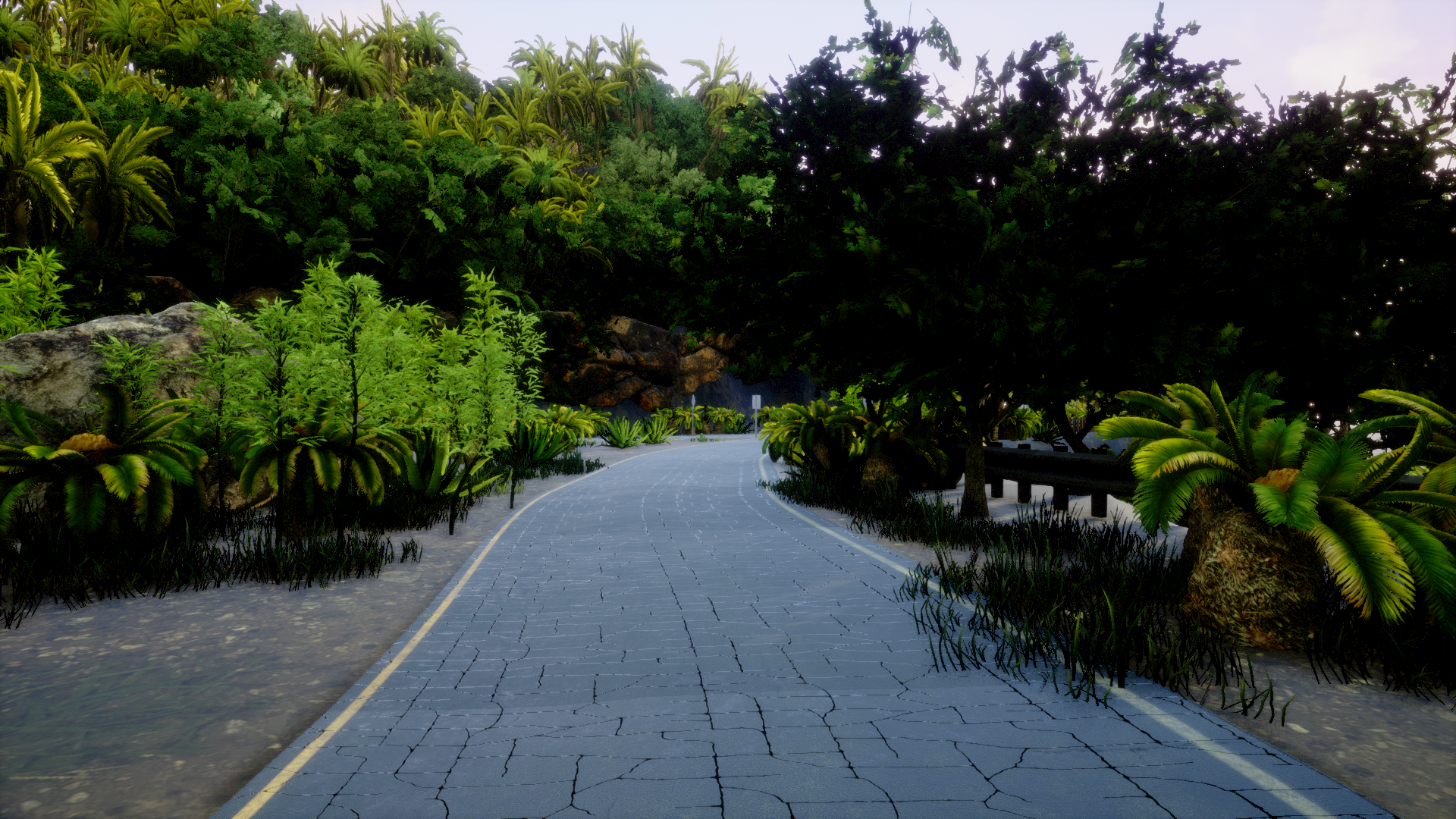}}
        \vspace{2pt}
    \end{minipage}
    \begin{minipage}{0.32\linewidth}
        \centerline{\includegraphics[width=\textwidth]{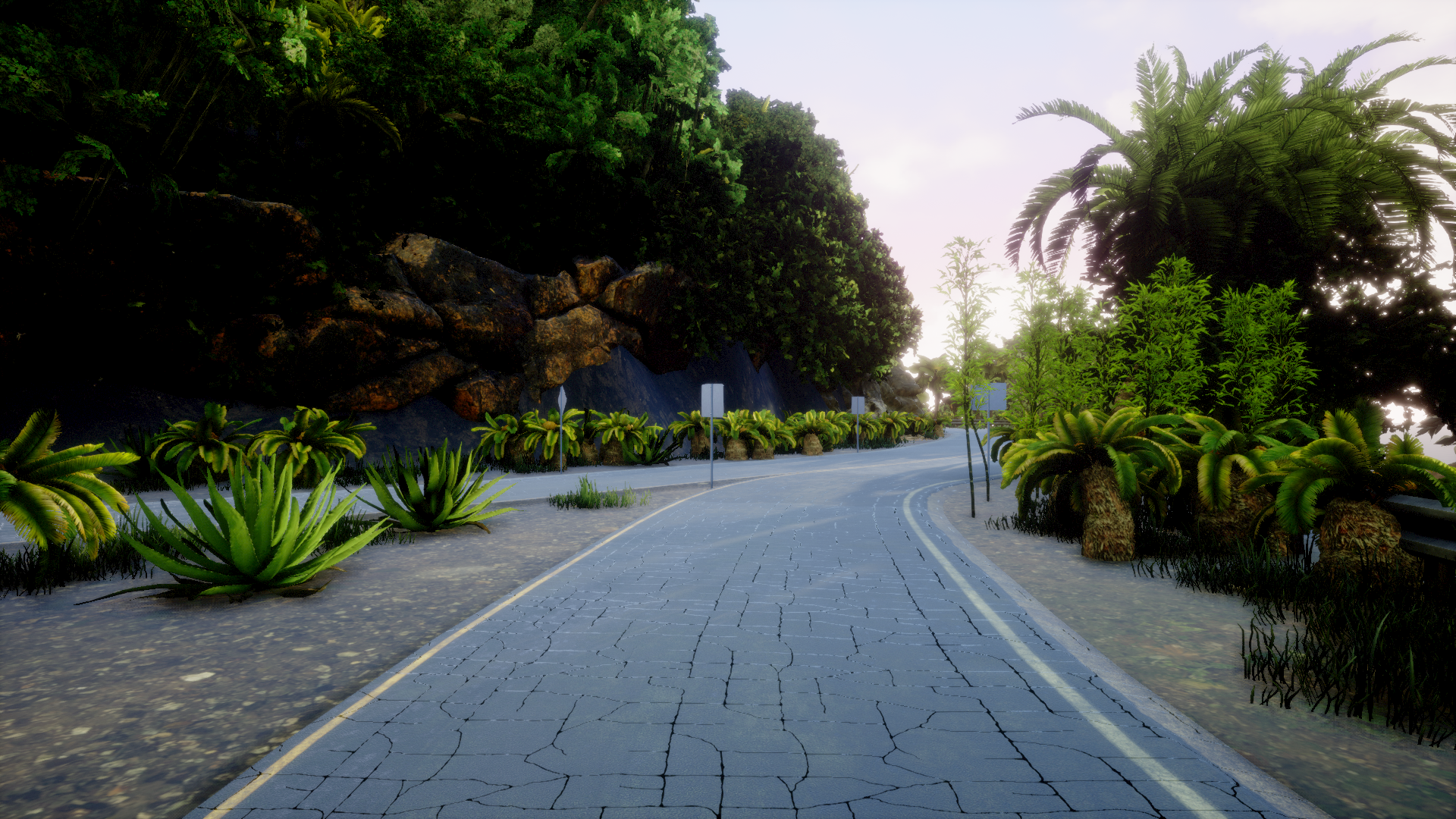}}
        \vspace{2pt}
    \end{minipage}
    \begin{minipage}{0.32\linewidth}
        \centerline{\includegraphics[width=\textwidth]{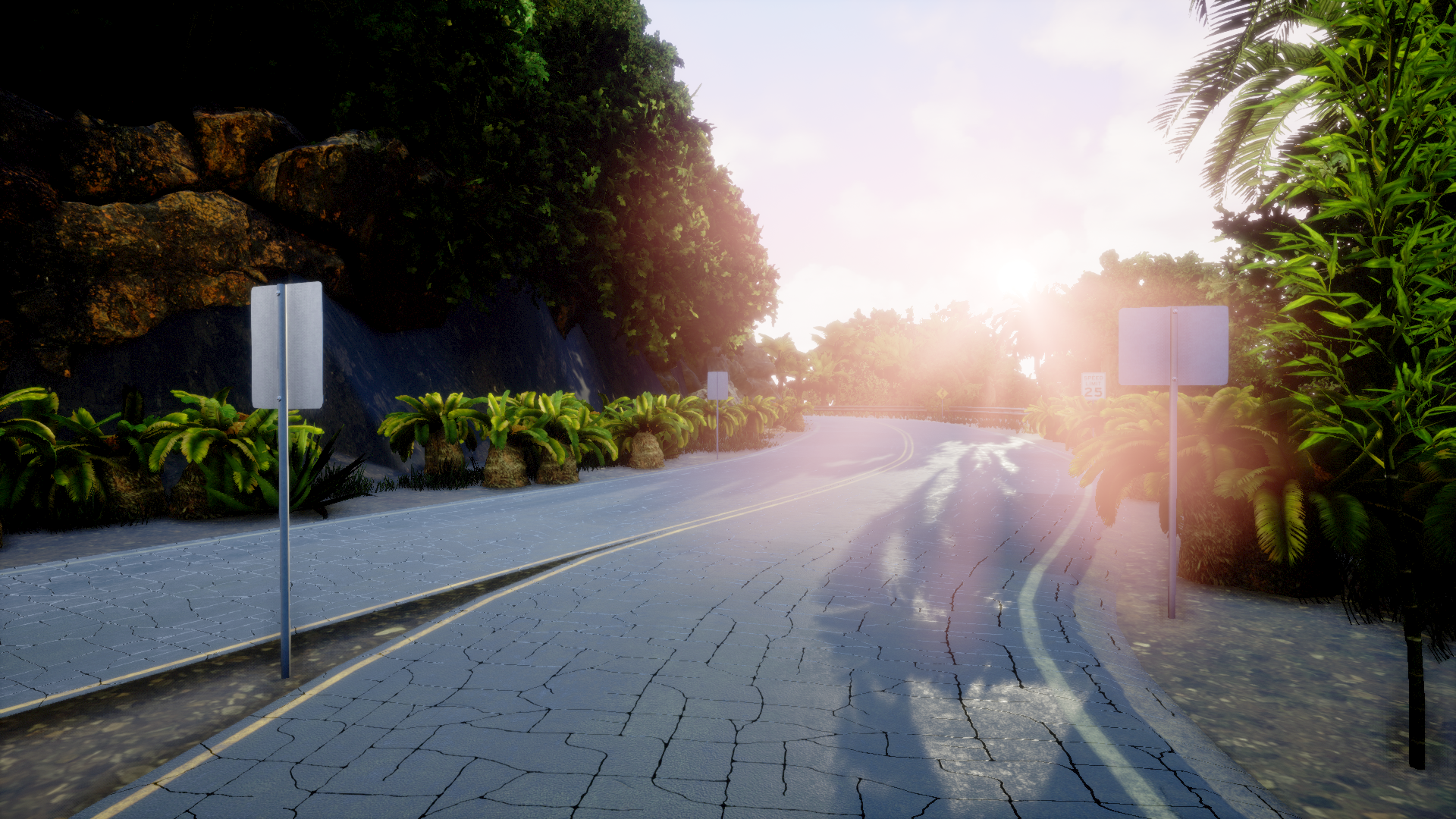}}
        \vspace{2pt}
    \end{minipage}
    \caption{The Used Environment (Coastline) in Our Simulation.}
    \label{fig:environ_example}
\end{figure}
\textbf{Hardware and Simulation Setup.} Our prototype as shown in in Fig. \ref{fig:codrone_implementation_overview} consists of:
\begin{itemize}
    \item An onboard Jetson Nano for drone computation.
    \item An edge server equipped with a 16-core 3.7GHz Intel CPU and an NVIDIA GeForce RTX 4090 GPU.
    \item Vision-Language Model: For visual reasoning and instruction following, we employ Qwen-VL-Max \cite{2023arXiv230916609B}, a state-of-the-art vision-language model developed by Tongyi Lab. It excels in complex visual comprehension tasks, enhancing the drone’s scene understanding and decision-making.
    \item Depth Estimation Model: During autonomous navigation, inference speed is critical for real-time performance. To balance accuracy and efficiency, we adopt Depth Anything V2 \cite{depth_anything_v2}, an efficient depth estimation model.

\end{itemize}
Experiments are conducted in AirSim \cite{shah2018airsim}, a high-fidelity UAV simulator built on Unreal Engine (UE). AirSim provides programmable image and control APIs, enabling rigorous testing of perception and navigation algorithms. The \textbf{CoDrone} system is implemented in Python, with neural networks trained using PyTorch. Additionally, we deploy a Tornado-based web server on the edge device to facilitate HTTP-based communication between modules, ensuring seamless data exchange.


\begin{figure*}
    \begin{minipage}{.32\linewidth}
        \includegraphics[width=\linewidth]{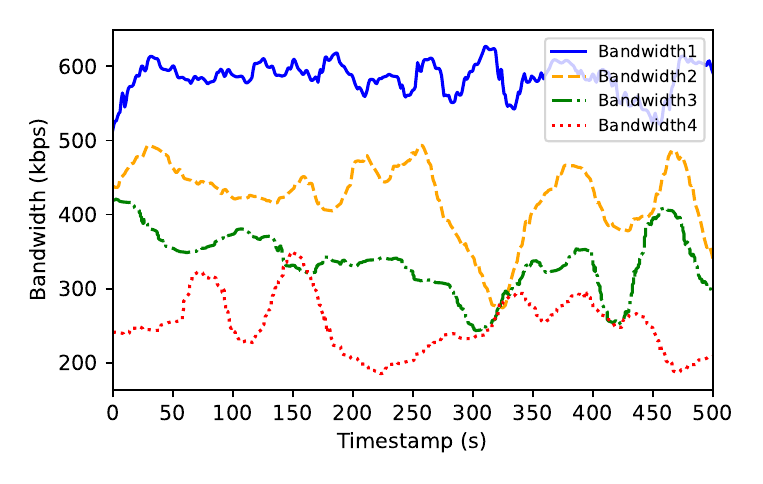}
        \caption{The used bandwidth traces in our experiments}
        \label{fig:bandwidth}
    \end{minipage}
    \begin{minipage}{.32\linewidth}
        \centering
        \includegraphics[width=\linewidth]{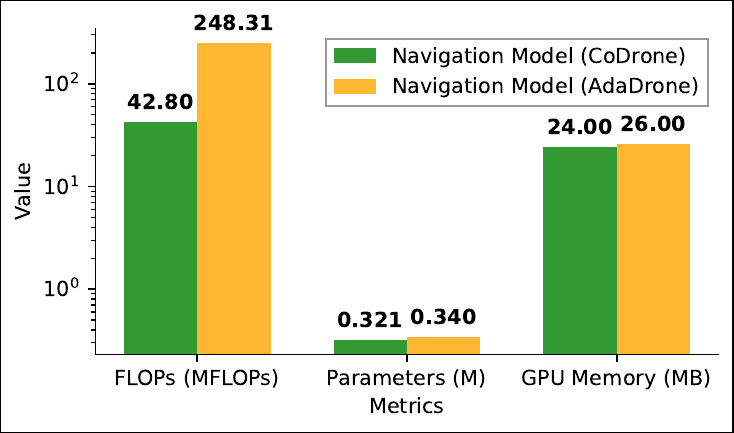}
        \caption{Performance Comparison: CoDrone vs AdaDrone model}
        \label{fig:res:nav_compare}
    \end{minipage}
    \begin{minipage}{.32\linewidth}
        \includegraphics[width=\linewidth]{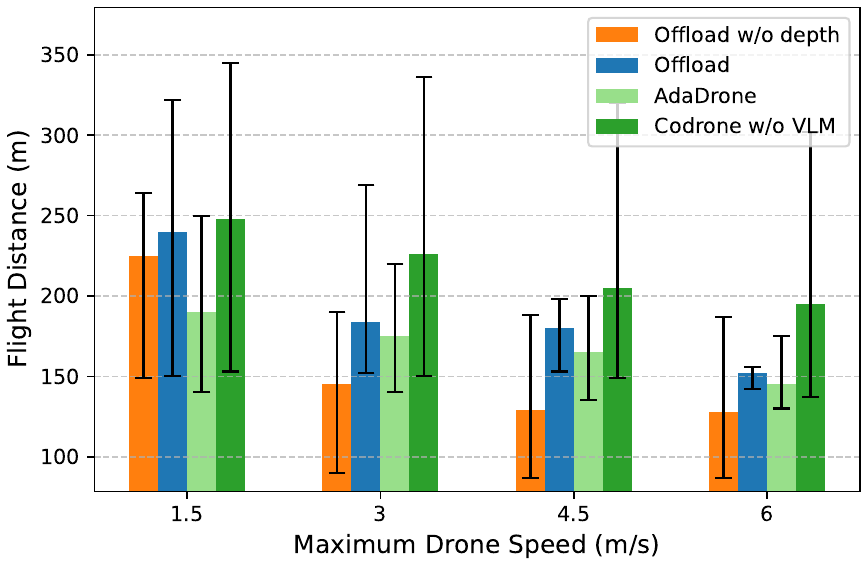}
        \caption{flight distance with various speed.}
        \label{fig:res:distance}
    \end{minipage}
\end{figure*}

\begin{figure*}
    \begin{minipage}{.32\linewidth}
        \includegraphics[width=\linewidth]{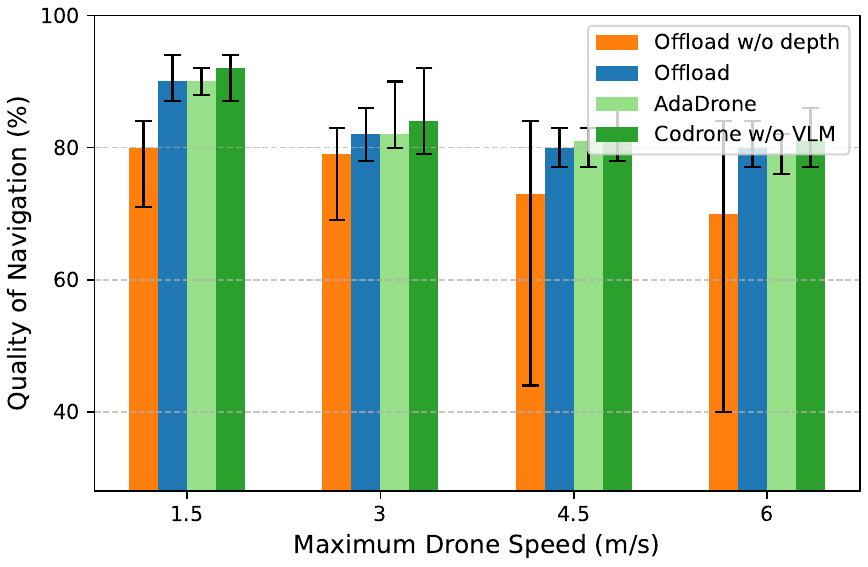}
        \caption{QON with various speed.}
        \label{fig:res:qon_distance}
    \end{minipage}
    \begin{minipage}{.32\linewidth}
        \centering
        \includegraphics[width=\linewidth]{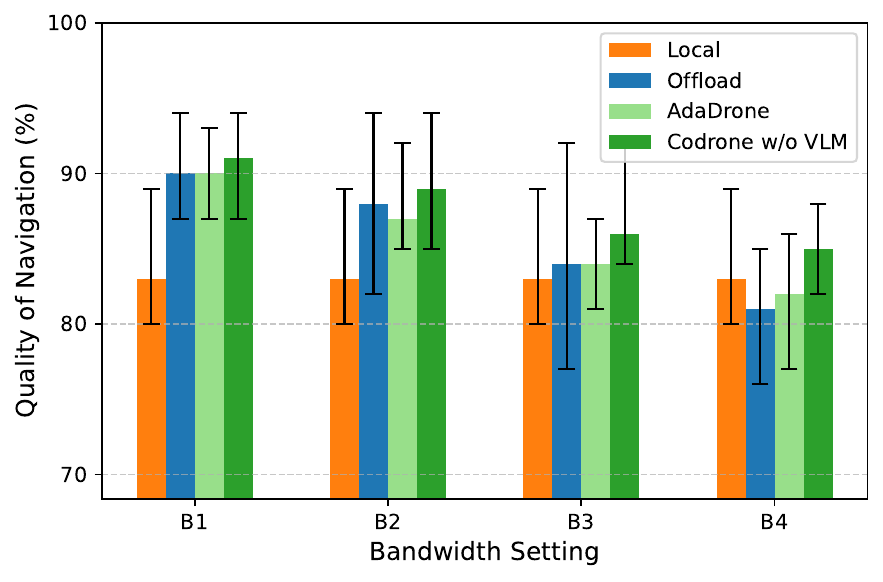}
        \caption{QON with various bandwidths.}
        \label{fig:res:qon_bandwidth}
    \end{minipage}
   \begin{minipage}{.32\linewidth}
        \includegraphics[width=\linewidth]{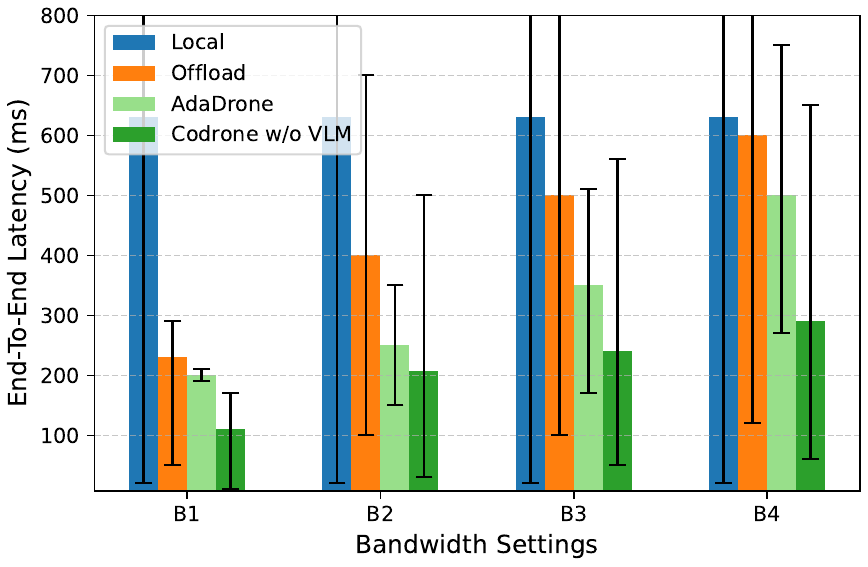}
        \caption{Latency with various bandwidth.}
        \label{fig:res:latency}
    \end{minipage}
\end{figure*}
\section{Performance Evaluation}
\label{sec:performance_eval}
We conduct extensive experiments to evaluate two key aspects of \textbf{CoDrone}: 1) The efficacy of the neural scheduler combined with depth-estimation-based occupancy grid maps in enabling robust autonomous navigation; 2) The impact of integrating a UAV-oriented VLM on handling complex, dynamic environments. Our experiments address the following research questions:

\textit{\textbf{Q1}}: How effectively does the neural scheduler, leveraging depth estimation for occupancy grid mapping, enable autonomous drone navigation across diverse environments?

\textit{\textbf{Q2}}: Can the neural scheduler maintain performance under dynamic edge network conditions (e.g., latency, bandwidth fluctuations)?

\textit{\textbf{Q3}}: To what extent does the vision-language model enhance obstacle avoidance and trajectory correction? 

\textit{\textbf{Q4}}: How does CoDrone’s integration of depth maps and VLM improve overall \textit{autonomy}?

\subsection{Methodology}
\textbf{Scenarios.} To evaluate CoDrone's environmental adaptability, we conduct experiments in AirSim \cite{shah2018airsim} using the ``Coastline" environment, which provides realistic coastal terrain features shown in Fig. \ref{fig:environ_example}. Additionally, we construct four bandwidth fluctuation scenarios derived from the HSDPA dataset \cite{10.1145/2483977.2483991} and impose controlled bandwidth constraints to more faithfully emulate the wireless communication conditions between the drone and the server. These four scenarios represent varying network conditions, including cases with relatively high bandwidth and low fluctuation, as well as relatively high bandwidth with significant fluctuations (see Fig. \ref{fig:bandwidth}). This comprehensive setup enables rigorous evaluation of system performance across diverse environmental and network conditions.


\textbf{Metrics.} To achieve an optimal trade-off between the autonomous navigation task and the inspection analysis task, we evaluate the effectiveness using the following metrics: for the autonomous navigation task, we focus on 1) \textit{Quality of Navigation (QoN)}. We use Eq. (\ref{metric:qon}) to calculate the QoN. The depth value represents the deviation from the front obstacles and the possible collision, which indicates that the QoN reflects the real effectiveness of autonomous navigation. 2) \textit{Flight distance}, a common and direct metric that reflects the total distance traversed from takeoff to safe landing or unplanned course deviation. To ensure robustness, we repeat the flight five times and average the recorded distances, mitigating potential environmental or operational variability. 3)\textit{End-to-End Latency}. In autonomous navigation, end-to-end latency measures the time from command issuance to full execution. Calculated as the time between command start and system stabilization, lower latency aids obstacle avoidance. We measure it across multiple test flights in a coastline environment. For each, we meticulously record the time of each inference and average to account for external variables, assessing real-time navigation ability.

\textbf{Baseline.} We design five heuristics as baseline strategies and compare CoDrone with them:
1) \textit{AdaDrone\cite{chen2022adadrone}}. This method treats adaptive navigation as a service and designs a DRL-based neural scheduler. It performs inference using RGB images and is capable of dynamically adapting the resolution, model execution placement, and image encoding quality in response to variations in environmental and network conditions. By jointly optimizing these factors, AdaDrone enables high-quality autonomous flight for drones.
2) \textit{CoDrone w/o VLM}.This method is the ablation baseline obtained by removing the VLM module in the cloud from the full CoDrone system while retaining all other components unchanged.
3) \textit{Local}.This method is a non-adaptive approach. It mandates the deployment of the Edge Navigation Module onto the onboard computing unit to ensure immediate execution at any given time. As for the depth estimation module, we incorporate a random number mechanism to allow a one-third probability of utilizing the depth estimation model for assisted navigation during each navigation inference. This stringent requirement implies that it is the most prevalent choice when the drone is capable of carrying a computing device with sufficient computational power.
4) \textit{Offload}.This approach adopts a non-adaptive strategy. It always gives priority to running the navigation model and the depth estimation model on the server. There is a one-third probability of using a fixed RGB image compression ratio of 95\% and relying on the depth estimation model for assistance and a two-thirds probability of using grayscale images with a fixed 95\% image compression ratio without the aid of the depth estimation model. Such a design makes this method a common choice when the drone lacks sufficient computational resources but can establish a stable network connection with the server. At the same time, this method never invokes the VLM module approach for flight assistance under any circumstances.
5) \textit{Offload w/o Depth Map}. This method is a baseline configuration in which both the depth estimation module in the edge and the VLM module in the cloud are disabled, and all lightweight navigation decisions are made solely by the edge server based on raw image input.

\subsection{Performance Gain in CoDrone w/o VLM}
In response to \textbf{\textit{Q1}}, we demonstrate the effectiveness of autonomous navigation with the assistance of a depth model by comparing it with three other methods. Fig.\ref{fig:res:nav_compare} compares the performance differences among various inference models.  Fig.\ref{fig:res:distance}, Fig.\ref{fig:res:qon_bandwidth} and Fig.\ref{fig:res:qon_distance} illustrate the flight distances of these methods in different scenarios under four preset bandwidth conditions.

For navigation models, regarding images with the same resolution of 200×200, since CoDrone performs inference using grayscale images, its GPU memory usage and model parameters are only 1\% lower than those of adaDrone, yet its inference computation load is less than one-third of that of adaDrone. According to flight distance, at any speed, the method with depth estimation, due to the assistance of the depth estimation model under specific circumstances, outperforms the other three methods in terms of the maximum, average, and minimum flight distances. Compared with the AdaDrone method, its average flight distance can be extended by up to 86 meters, and the maximum flight distance can be extended by up to 173 meters. To further explore the role of the depth estimation model, we divided the offloading method into two categories: one that uses the depth estimation model (offload) and one that does not (offload-w/o depth). From the results, it can be seen that when the speed increases, the minimum distance of offload w/o depth drops rapidly. This may be because it fails to decelerate well in some complex environments, leading to collisions. However, the offload method can obtain the distance of current obstacles from the depth map, enabling it to better cope with complex environments. Therefore, at different speeds, whether it is the maximum, average, or minimum value, it is much higher than the offload w/o depth method, and the flight distance can be extended by up to 80 meters. From the perspective of QoN, CoDrone consistently outperforms other baselines, achieving a maximum improvement of up to 7\% in navigation quality.

The performance improvement of the method assisted by depth estimation is attributed to its ability to adaptively and promptly incorporate the information of an impending collision into the control information and transmit it back to the drone before a collision occurs.

\subsection{Adaptability}

\begin{figure}[htbp]
    \centering
    \includegraphics[width=.98\linewidth]{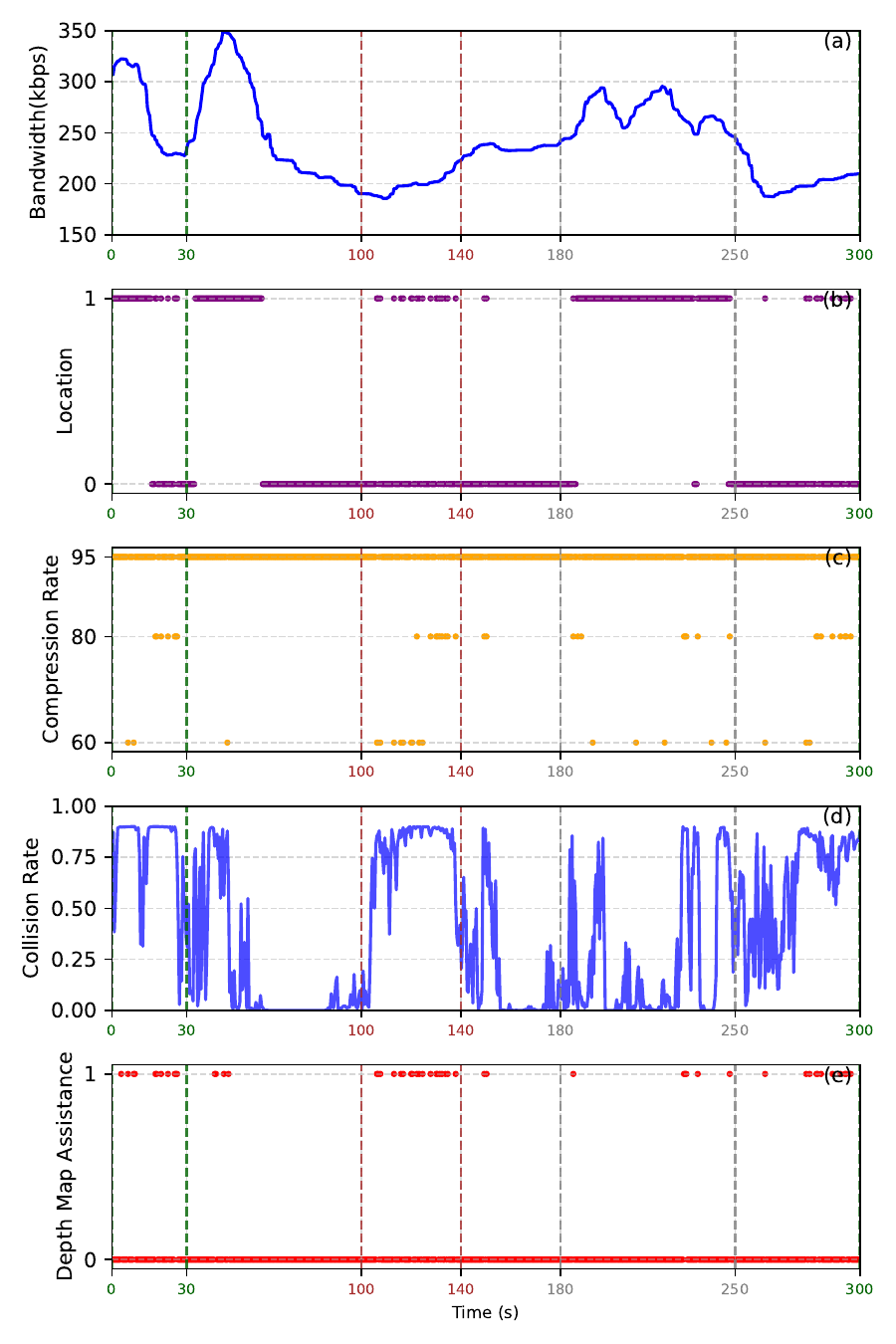}
    \caption{Case study of CoDrone’s adaptive decisions. (a) Bandwidth trace.(b) Decided execution locations of navigation model inference, where 0 and 1 stand for onboard device and server, respectively.  (c) CoDrone’s compression ratio decisions (d) the collision rate inference by navigation model. (e) the chosen depth estimation model for navigation assistance, where '1' indicates active use and '0' indicates non-use.}
    \label{fig:res:case_study}
\end{figure}

\begin{figure*}[htbp]
    \centering
    \begin{minipage}{0.99\linewidth}
        \centerline{\includegraphics[width=\textwidth]{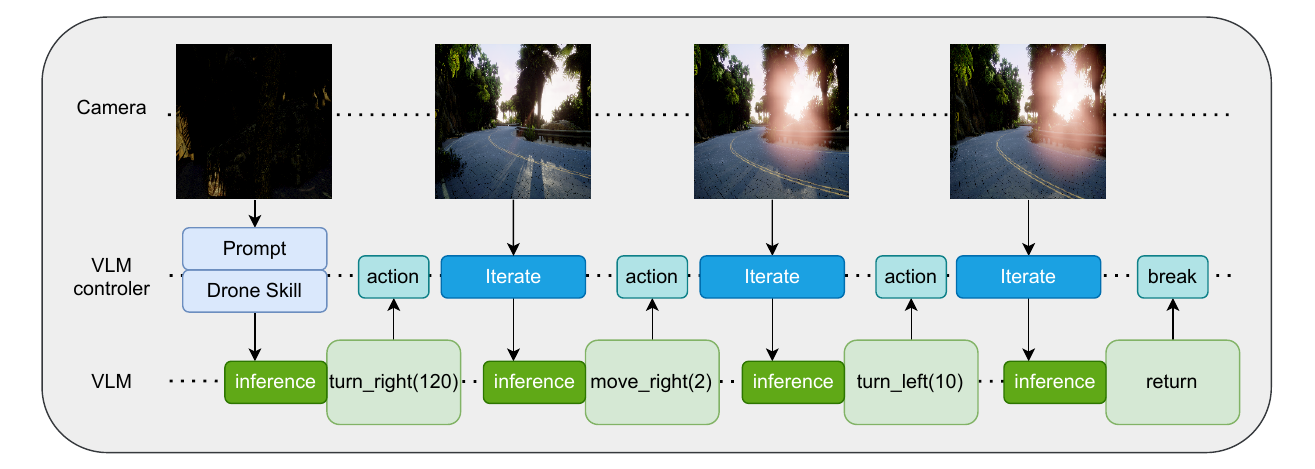}}
        \caption{VLM-assisted flight in complex environments.}
        \vspace{2pt}
    \end{minipage}
\end{figure*}
 \begin{figure*}[htbp]
       \begin{minipage}{.48\linewidth}
        \includegraphics[width=\linewidth, height=5cm]{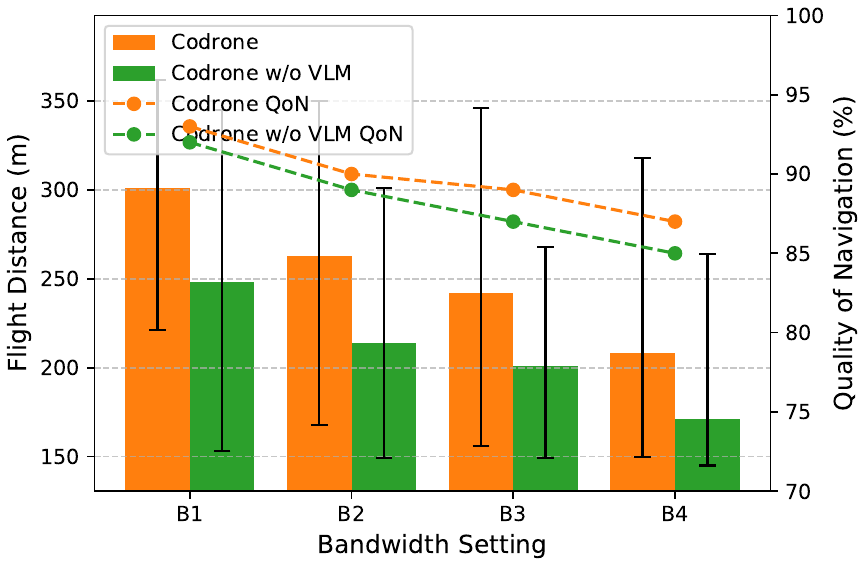}
        \caption{Quality of Navigation and distance with various bandwidth.}
        \label{fig:res:llm-bandwidth}
    \end{minipage}
   \begin{minipage}{.48\linewidth}
        \includegraphics[width=\linewidth, height=5cm]{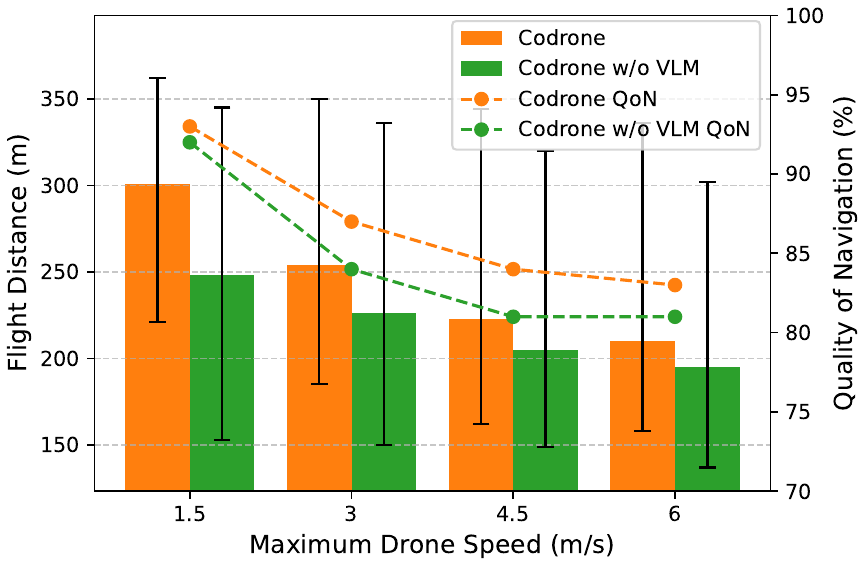}
        \caption{Quality of Navigation and distance with various speeds.}
        \label{fig:res:llm-speed}
    \end{minipage}
\end{figure*}
To answer \textbf{\textit{Q2}}, Firstly, we investigate the End-To-End Latency that the CoDrone and baseline methods achieved within the flights, and the results are illustrated in Fig. \ref{fig:res:latency}. Each figure independently demonstrates CoDrone's adaptive performance across varying bandwidth settings.
As shown in the figure, CoDrone achieves significantly lower end-to-end latency compared to both the end-to-end-only and offloading-based methods. Compared to AdaDrone, CoDrone incurs an additional inference step due to the integration of the depth estimation model. Moreover, as this model requires RGB input rather than grayscale imagery, the associated transmission delay is increased, leading to a higher maximum end-to-end latency. However, thanks to the effective scheduling strategy of the Neural Scheduler Module and the substantial reduction in computational load of the navigation model enabled by grayscale image processing, CoDrone achieves lower average and minimum latencies across all four evaluated bandwidth conditions compared to AdaDrone.

Then, we perform a 300s flight in simulation in bandwidth 4.
We manually adjust the steering command of the drone to the right direction when it deviates. 
Fig. \ref{fig:res:case_study} shows the details of the environment and the chosen configuration of CoDrone throughout the flight. 
In the initial phase of the flight (0-30 seconds), when the drone's navigation model determines that the current collision rate is high, CoDrone will use the depth estimation model for auxiliary navigation to achieve better flight performance.
In response to the significant bandwidth drop between 80 and 120 seconds shown in the figure, CoDrone adaptively adjusts by performing local inference for navigation or increasing the compression rate of image transmission, thereby enabling faster data transfer and obtaining inference results more quickly.
Meanwhile, during the period from 100 to 140 seconds, the collision rate rises over a certain distance. Based on the current collision rate, CoDrone uses the depth estimation model for auxiliary navigation to a certain extent.
As the bandwidth increases and the complexity and dynamics of the environment decrease (180 - 250 seconds), CoDrone then conducts part of the inference at the edge based on the current bandwidth. Finally, as the bandwidth drops between 250 and 300 seconds, CoDrone opts to perform the navigation inference on board. 

These two experiments demonstrate that the Neural Scheduler Module is capable of effectively orchestrating task offloading and adaptively invoking auxiliary models in response to dynamically changing environmental and network conditions.

\subsection{Performance of Vision-Language Models in Complex Environments}
To answer \textbf{\textit{Q3}}, we logged the use of the vision-language Model Module during drone flight to evaluate the impact of the VLM on autonomous cruise flight. During the experiments, the UAV encountered various extreme scenarios. For instance, it inadvertently flew into bushes alongside the road. In response, CoDrone fed the front-camera image to the VLM, which determined that the UAV had deviated from its intended path and invoked the \texttt{turn\_right} function with a parameter of 120, rotating the UAV 120 degrees clockwise.
Following this maneuver, the VLM received the updated image and historical data, identifying the UAV as being near the roadside. It then called \texttt{move\_right(2)} to shift the UAV 2 meters to the right. Upon further evaluation, the VLM detected that the UAV was still not aligned with the road center and invoked \texttt{turn\_left(10)} to rotate it 10 degrees counterclockwise. After this adjustment, the VLM confirmed that the UAV had returned to the correct position and orientation, ceased issuing commands, and allowed the UAV to continue its flight.

This case study illustrates the VLM's capability to effectively interpret visual and positional data to assess the UAV’s state and appropriately utilize control functions to avoid obstacles and realign with the intended flight path.

\subsection{Performance Gain in CoDrone}
To address \textbf{\textit{Q4}}, we conducted targeted ablation studies. Specifically, to highlight the contribution of the vision-language model, our ablation counterpart, when encountering scenarios that typically engage large language models (LLMs), would suspend in mid-air without translation for a predetermined duration before resuming navigation with the primary navigation model.
Fig.\ref{fig:res:llm-bandwidth} illustrates that, under scenarios with identical maximum speed (1.5 m/s) but varying network bandwidths, the CoDrone integrated with a vision language model consistently outperforms its ablation counterpart in terms of the maximum, minimum, and average flight distances. The maximum average difference in flight distance reached 62 meters, and the average QoN differed by up to 3\%. Fig.\ref{fig:res:llm-speed} further demonstrates that, under similar conditions but with varying maximum speeds, CoDrone consistently achieves higher average flight distances compared to its ablation counterpart. The maximum average flight distance difference between the two approaches was observed to be 53 meters, with an average QoN improvement of up to 3\%.
These findings suggest that the vision-language model significantly assists the UAV in obstacle avoidance and path adherence when navigating complex environments. The synergistic integration of vision-language models with traditional navigation methodologies yields enhanced performance.

\section{Conclusion}
\label{sec:conclusion}
In this paper, we propose CoDrone, a foundation model-enabled edge computing framework for autonomous UAV navigation, which facilitates efficient collaboration among cloud, edge, and end devices to ensure safe and reliable flight operations. By leveraging grayscale imagery, the framework significantly reduces data volume while preserving essential environmental information, thereby lowering both computational latency and communication overhead. It further incorporates a depth estimation module to augment environmental perception and leverages a VLM for semantic scene understanding, thereby enhancing the system’s robustness and adaptability under dynamic operational conditions. Additionally, CoDrone incorporates an invocation strategy and a DRL-based neural scheduler to dynamically orchestrate task offloading and model activation in response to varying environmental and network conditions. Experimental results demonstrate that CoDrone outperforms existing baseline approaches in terms of flight distance, end-to-end latency, and QoN, highlighting its superiority in dynamic and resource-constrained operational environments.

\bibliographystyle{IEEEtran}
\bibliography{bib/export.bib}

\end{document}